\documentclass[Afour,sageh,times]{sagej} 

\usepackage{moreverb,url}

\usepackage[colorlinks,bookmarksopen,bookmarksnumbered,citecolor=red,urlcolor=red]{hyperref}

\usepackage{styles/basic,styles/math,styles/figures,styles/tables,styles/algorithms}

\newcommand\BibTeX{{\rmfamily B\kern-.05em \textsc{i\kern-.025em b}\kern-.08em
T\kern-.1667em\lower.7ex\hbox{E}\kern-.125emX}}

\def\volumeyear{2024}

\begin{document}

\runninghead{Fichera and Billard}

\title{Learning Dynamical Systems Encoding Non-Linearity within Space Curvature}

\author{Bernardo Fichera\affilnum{1} and Aude Billard\affilnum{1}}

\affiliation{\affilnum{1}\'Ecole Polytechnique F\'ed\'erale de Lausanne, Lausanne, Switzerland}

\corrauth{Bernardo Fichera, Learning Algorithms and Systems Laboratory,
    \'Ecole Polytechnique F\'ed\'erale de Lausanne,
    Lausanne,
    Switzerland.}

\email{bernardo.fichera@epfl.ch}

\begin{abstract}
    Dynamical Systems (DS) are an effective and powerful means of shaping high-level policies for robotics control.
    They provide robust and reactive control while ensuring the stability of the driving vector field.
    The increasing complexity of real-world scenarios necessitates DS with a higher degree of non-linearity, along with the ability to adapt to potential changes in environmental conditions, such as obstacles.
    Current learning strategies for DSs often involve a trade-off, sacrificing either stability guarantees or offline computational efficiency in order to enhance the capabilities of the learned DS.
    Online local adaptation to environmental changes is either not taken into consideration or treated as a separate problem.
    In this paper, our objective is to introduce a method that enhances the complexity of the learned DS without compromising efficiency during training or stability guarantees.
    Furthermore, we aim to provide a unified approach for seamlessly integrating the initially learned DS's non-linearity with any local non-linearities that may arise due to changes in the environment.
    We propose a geometrical approach to learn asymptotically stable non-linear DS for robotics control.
    Each DS is modeled as a harmonic damped oscillator on a latent manifold.
    By learning the manifold's Euclidean embedded representation, our approach encodes the non-linearity of the DS within the curvature of the space.
    Having an explicit embedded representation of the manifold allows us to showcase obstacle avoidance by directly inducing local deformations of the space.
    We demonstrate the effectiveness of our methodology through two scenarios: first, the 2D learning of synthetic vector fields, and second, the learning of 3D robotic end-effector motions in real-world~settings.
\end{abstract}


\maketitle


\section{Introduction}
\label{sec:introduction}

Learning from Demonstration (LfD) represents a powerful approach to derive global behavioral policies for high-level closed-loop control by observing demonstrated tasks.
Such policies are represented using the mathematical framework of Dynamical Systems (DS), namely a vector field $\mathbf{f}: \mathbb{R}^d \rightarrow \mathbb{R}^d$, mapping the $d$-dimensional input state $\mathbf{x}(t) \in \mathbb{R}^d$ to its time-derivative $\dot{\mathbf{x}}(t) \in \mathbb{R}^d$, such that $\dot{\mathbf{x}}(t) = \mathbf{f}(\mathbf{x})$.

In the field of robotics, this framework is commonly employed for describing and regulating various motion types, such as point-to-point motions characterized by fixed-point stable equilibrium DS, or periodic motions featuring stable limit-cycle DS.
The stability of a learned DS becomes a significant concern when it is applied in closed-loop control systems. Using standard regression methods to learn the mapping $\mathbf{f}$ provides no inherent guarantee of producing stable control policies.
A wealth of learning approaches have been developed in the last decades to learn a DS with the stability guaranteed.
They follow two fundamental paths: 1) constraint optimization; 2) learning of complex potential function via diffeomorphism.

\begin{figure}[t]
    \centering
    \includegraphics[width=.4\textwidth]{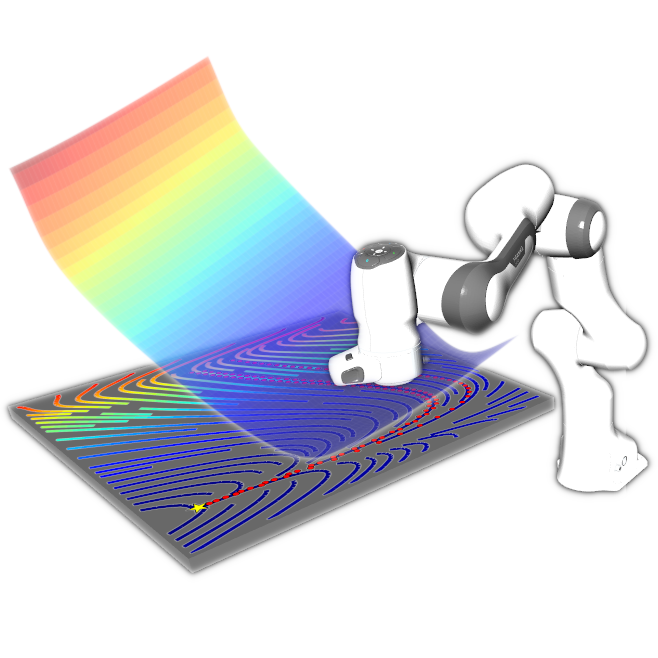}
    \caption{\footnotesize End-effector motion of a robotic arm guided by a 2D learned DS. The surface corresponds to the 3D Euclidean space embedded representation of the learnt 2D manifold. The color gradient represents the value of the potential function that drives the linear vector field taking place on the 2D manifold. The manifold's curvature induces "apparent" non-linearity in the 2D chart Euclidean space representation of the vector field taking place on the 2D manifold. During the learning process, the curvature of the manifold adapts so that the streamlines of the 2D chart Euclidean space representation of the vector field follow closely the demonstrated trajectories (red dots), preserving the stability towards a desired equilibrium point (yellow star).}
    \label{fig:motion_example}
\end{figure}
In the first category, the most popular approach is to derive constraints via Lyapunov's second method for stability.
In the beginning, \cite{khansari-zadeh_learning_2011} utilized a quadratic Lyapunov function to establish stability conditions in a Gaussian Mixture Regression (GMR) problem for learning Dynamical Systems (DS).
However, stability guarantee imposes a severe restriction on the learnable complexity of the DS and prevents learning highly non-linear DS containing high-curvature regions or non-monotonic motions (i.e., temporally moving away from the attractor).
More recent approaches tried to alleviate this issue by improving the complexity of the Lyapunov function adopted as constraint, \cite{figueroa_physicallyconsistent_2018}.
Specifically, by moving towards an elliptic Lyapunov function, these approaches are capable of relaxing the constraints allowing for learning more complicated trajectories.
Nevertheless, they still struggle in learning DS exhibiting high non-linearity and non-monotonic behavior in different radial directions with respect to the equilibrium point.

An alternative constraint optimization problem can be derived from Contraction Theory (CT), \cite{lohmiller_contraction_1998}.
Abstracting from the absolute position of the equilibrium point, CT follows a differential perspective.
Conditions derived by CT impose local contraction of trajectories implying, as a consequence, global exponential stability towards the equilibrium point.
\cite{blocher_learning_2017} takes advantage of CT to derive a stabilizing controller that eliminates potential spurious attractor present in the DS learned without stability constraints.
\cite{sindhwani_learning_2018} uses CT to derive constraints for learning DS in a Support Vector Regression problem.
Both \cite{blocher_learning_2017} and \cite{sindhwani_learning_2018} rely on non-generalized contraction analysis, which, in turn, results in overly conservative constraints.
This is analogous to the adoption of a simplistic quadratic function in the Lyapunov approach to the stability problem.
In their work, \cite{ravichandar_learning_2017} introduced a GMR-based regression problem, incorporating stability constraints derived from generalized (CT) analysis.
While this approach demonstrates superior performance by relaxing overly conservative constraints, it does so at the expense of achieving global stability, focusing solely on local stability.

Another approach to learning DS involves the existence of a \emph{latent} space, in which either the Lyapunov function is quadratic or the DS is linear.
In these approaches, the focus is on learning a diffeomorphism $\psi: \mathbb{R}^d \rightarrow \mathbb{R}^d$ between the original space and the latent one.
A first example of this approach for solving the stability vs accuracy dilemma was proposed by \cite{neumann_learning_2015}.
This approach extends the applicability of SEDS by introducing a diffeomorphic mapping that transforms non-quadratic Lyapunov functions, ensuring point-wise stability of the demonstrated trajectory, into quadratic forms.
After applying SEDS in this transformed space, the desired policy is obtained through the application of the inverse diffeomorphic mapping.
More recent approaches concentrate on directly identifying latent spaces where the DS exhibits linear behavior.
These methods make use of an approximation of the Large Deformation Diffeomorphic Metric Mapping (LDDMM) framework, \cite{joshi2000landmark}, to accommodate the required smoothness constraints in the mapping.

In \cite{perrin_fast_2016}, the diffeomorphic learning DS is fundamentally geometric, focusing solely on the positions of the original and target points. To reconstruct the proper velocity profile, a rescaling of the learned DS is employed. The diffeomorphic map is learned as sequence locally weighted translations applied to the points in the original space. Additionally, more modern and network-based strategies, such as non-volume preserving transformation (NVP) \cite{dinh_density_2016}, can be utilized to model the diffeomorphic map.
\cite{rana_euclideanizing_2020} adopts NVP transformations within an optimization framework that incorporates dynamic information, specifically velocity, into the process.
This results in a one-step learning algorithm.
Essentially, all these approaches involve learning a complex potential function whose gradient closely follows the target DS.
While these methods demonstrate improved performance, they come with the trade-off of requiring sophisticated machinery for creating a function approximator capable of learning a mapping that exhibits the diffeomorphic property mandated by the proposed mathematical framework.

All the methods discussed above are confined to learning first-order conservative DS. Dissipative or second-order DS cannot be learned within this framework.
Moreover, online local adaptation to environmental changes is not considered as part of the problem.
In DS-based control, such issues are typically addressed a-posteriori and handled through a modulation matrix, as in \cite{khansari-zadeh_dynamical_2012}.
These approaches are agnostic to the DS they aim to modulate, potentially leading to spurious attractors whenever the DS velocity direction aligns with the normal principal direction of the modulation matrix.
A clever trick to partially address this problem involves breaking orthogonality between the modulation matrix's principal components, as proposed by \cite{huber_avoidance_2019}.
However, these methods rely on manually designing modulation matrices for each local adaptation they aim to accommodate, resulting in increased complexity in both problem design and stability analysis.
The application of modulation to second-order systems remains unclear.

These limitations cast a shadow over DS methods when compared to planning methods.
Planning methods, armed with inherent adaptability and increasingly efficient sampling-based strategies, \cite{williams_model_2017}, are gradually overcoming their reactivity challenges, fueled by advancements in computational hardware power, as shown in \cite{bhardwaj_storm_2021}.
In response to this, geometry-based DS shaping approaches, drawing on tools from the field of differential geometry, emerge as a solution to reverse this trend.
These approaches aim to achieve two crucial objectives: 1) enhance DS policies with greater expressivity and bolster their adaptability, and 2) broaden the modularity of the approach to tackle the growing complexity of real-world scenarios in which robotic systems must operate.

Drawing inspiration from \cite{bullo_geometric_2005}, \cite{ratliff_riemannian_2018} introduced the Riemannian Motion Policy (RMP), a modular mathematical framework for robotic motion generation.
In contrast to prior works, this approach produces second-order DS, the specific behavior of which is inherently linked to a Riemannian metric.
By carefully designing such metrics, a wide range of behaviors can be exhibited and combined.

With few exceptions, detailed in \Cref{sec:related}, this research line has not explicitly focused on Learning from Demonstration (LfD).
Instead, the emphasis has been on expanding the capabilities of the mathematical framework to enhance the complexity and variety of reproducible behaviors.
Building upon RMP, \cite{cheng_rmpflow_2020} proposed RMPflow, which effectively combines different tasks designed via RMPs, leveraging the sparsity of the structure for computational efficiency.

Summarizing the endeavors of previous works, \cite{bylard_composable_2021} provides a principled and geometrically consistent description of the mathematical framework used for geometry-based policies.
Current research in the field is shifting towards a more general formalism, extending beyond Riemannian differentiable manifolds to include Finsler structures, as discussed in \cite{xie_geometric_2021, ratliff_generalized_2021}.
This expansion aims to introduce velocities as a fundamental ingredient in shaping metrics that define policies' behavior.

\subsection*{Contribution} \label{sec:contribution}

Our work bridges the gap between DS learning literature and the evolving field of geometry-based shaping of DS policies.
From the DS learning literature, we draw inspiration from concepts related to the existence of a latent space.
A notable distinction is that we do not enforce diffeomorphic constraints.
On the other hand, we borrow from the Geometric DS literature the idea of a chart-based representation of DS occurring on a manifold, along with employing various tools from differential geometry to define the operators we utilize.

In this work, we introduce a novel approach to learning DS that aims to integrate LfD with modern geometric control techniques.
Within our framework, the non-linearity of the DS is "encoded" within the curvature of a $d+1$-dimensional latent manifold, where $d$ represents the dimension of the vector field being learned.
This concept is illustrated in~\Cref{fig:motion_example}.

Our framework naturally extends to second-order dissipative DS and easily adapts to potential online local non-linearity changes, such as those arising from the presence of obstacles.
Additionally, we propose a variety of solutions, such as directional and exponential, to make the usage second-order DS effective in LfD scenario.
The proposed geometric framework, as indicated by standard metrics for evaluating learning performance, matches or outperforms the state-of-the-art while achieving notably lower computational costs during both training and query phases.
These efficiency gains are obtained without compromising the performance or stability of the learned DS.
Furthermore, such framework amplifies the expressivity of geometrical policies, shedding light on the relationship between DS non-linearity and manifold curvature.
It also provides an explicit visualization of the Euclidean embedded representation of the latent manifold responsible for generating non-linearity.

To operationalize this work, we developed a fully differentiable \textsf{PyTorch} library\footnote{\textsf{learn-embedding} code available at:\\\url{https://github.com/nash169/learn-embedding}}, which can be used for both Learning from Demonstration (LfD) and manually shaping geometric policies. In second-order settings, such policies can exhibit either geodesic or damped harmonic behavior, expanding the variety of behaviors available.
To preserve reactive control features, we developed a high-performance \textsf{C++} library\footnote{\textsf{control-lib} code available at:\\\url{https://github.com/nash169/control-lib}}, that integrates our geometrical DS with fast one-step model-based or model-free Quadratic Programming control techniques.
Additionally, the fully-templated nature of the library allows for the generalization of the controllers' suite to different non-Euclidean spaces, such as Lie Groups.
In practical terms, for robot end-effector control, this represents a valuable feature for $\mathbb{R}^3 \times \text{SO}(3)$ control, where one controller operates in the three-dimensional Euclidean space, while the other one functions in the Special Orthogonal Group characterizing orientation space.
The control strategy is fully modular and easily integrable in RMP frameworks as in \cite{fichera_hybrid_2023}.

\section{Related work} \label{sec:related}

Our method differs from \cite{rana_euclideanizing_2020} by avoiding the need to learn a diffeomorphism between two chart representations of a manifold.
Instead, we focus on learning an embedding for a higher-dimensional Euclidean representation, which is computationally efficient and offers more expressive models.
Our approach is less mathematically restrictive, requiring only a homeomorphic relationship between the manifold and its embedding, and can use any continuous function approximator, bypassing the need for specific Non-Volume Preserving (NVP) networks.
This leads to a simpler and quicker optimization process.
Additionally, our method allows to construct second-order DSs, while still accommodating first-order DSs.
This flexibility enhances our model's ability to handle complex tasks, such as replicating crossing trajectories and navigating around concave obstacles using a combination of geodesic and harmonic motions.

While not directly classified as part of the LfD literature, \cite{mukadam_riemannian_2019} applies learning methods in RMPflow by introducing weight functions that hierarchically modify the Lyapunov functions associated with subtasks.
This enhancement improves the overall performance of the combined policy.
Contrastingly, our approach primarily focuses on learning the curvature of the manifold.
This enables the generation of complex policies without incurring in stability problems caused by the lack of convexity of the Lyapunov function.

In the work presented in \cite{rana_learning_2019}, the emphasis lies in learning the Cholesky decomposition of the metric tensor.
When compared with our approach, this method not only exhibits limitations by solely addressing first-order geometry but also mandates specifically designed and more complex function approximators to learn the lower diagonal matrix of the metric tensor Cholesky decomposition.
This structure limits the flexibility to dynamically adjust the local geometry of the space to handle scenarios, such as obstacle avoidance.

\cite{beik-mohammadi_learning_2021} developed a method to generate geodesic motions aligned with observed trajectories by deriving a Riemannian metric from a map between the original and a latent space, learned through a Deep Autoencoder.
This involves geodesic motions generated via an iterative optimization process that minimizes the curve length between two points.
Conversely, our method focuses on constructing a DS formulation, ensuring stability, reactivity, and robustness against spatial and temporal disturbances typical in DS-based control.
We directly access the manifold as a higher-dimensional Euclidean representation by approximating a single embedding component through a simple feedforward network, eliminating the need for complex Deep Learning structures.
This not only enhances computational efficiency but also provides much greater expressivity, allowing a deeper understanding of why and how altering manifold's curvature to achieve the desired DS non-linearity.
Our framework allows for a directly modifiable embedding designed to facilitate online local deformation.


\section{Background}
\label{sec:background}
The presentation of the work relies heavily on concepts from differential geometry. Our notation follows \cite{carmo_riemannian_1992}. We employ the Einstein summation convention in which repeated indices are implicitly summed over.

Given a set, $\mathcal{M}$, and a Hausdorff and second-countable topology, $\mathcal{O}$, a topological space $(\mathcal{M}, \mathcal{O})$ is called a \emph{d-dimensional} manifold if \( \forall p \in \mathcal{M} : \exists \mathcal{U} \in \mathcal{O} : \exists x : \mathcal{U} \rightarrow x(\mathcal{U}) \subseteq \mathbb{R}^d \), with $x$ and $x^{-1}$  continuous maps.
$(\mathcal{U},x)$ is a \emph{chart} of the manifold $(\mathcal{M}, \mathcal{O})$.
$x$ is called the \emph{chart map}; it maps $p \in \mathcal{M}$ to the point $x(p) = \left( x^1(p), \dots, x^d(p) \right)$ into the $\mathbb{R}^d$ Euclidean space.
$\left( x^1(p), \dots, x^d(p) \right)$ are known as the coordinate maps or local coordinates.
With slight abuse of notation, we will refer to a point in $\mathbb{R}^d$ using the bold vector notation $\mathbf{x}=x(p)$, dropping the explicit dependence on $p \in \mathcal{M}$.
$x^i$ will be $i$-th local coordinate of $\v{x} \in \mathbb{R}^d$.

Throughout, we will denote with $\mathcal{M}$ a \emph{differentiable Riemannian} manifold, that is a manifold endowed with a $C^{\infty}$-atlas, $\mathcal{A}$, and a $(0,2)$-tensor field, $g$, with positive \emph{signature}, satisfying symmetry and non-degeneracy properties. We refer to $g$ as a Riemannian \emph{metric}.

$T_p\mathcal{M}$ ($T^*_p\mathcal{M}$) denotes the tangent (resp. cotangent) space at $p \in \mathcal{M}$.
We denote by $v_p \in T_p\mathcal{M}$ a vector in the tangent space at $p$.
Given a set of local coordinates $(x^1,\dots , x^d)$, in a neighborhood $\mathcal{U}$ of $p \in \mathcal{M}$, we denote by $\frac{\partial}{\partial x^i}$ (resp. $dx^i$) the $i$-th basis vector of $T_p\mathcal{M}$ (resp. $T^*_p\mathcal{M}$).
The tangent bundle $T\mathcal{M}$ (resp. cotangent bundle $T^*\mathcal{M}$) is the disjoint union of these tangent (resp. cotangent) spaces over all $p \in \mathcal{M}$.

A vector field (resp. covector field) $X$ on $\mathcal{U} \subset \mathcal{M}$ is a map assigning to each point $p \in \mathcal{U}$ a vector $X(p) \in T_p\mathcal{M}$ (resp. $X(p) \in T^*_p\mathcal{M}$).
$\Gamma(T\mathcal{M})$ (resp. $\Gamma(T^*\mathcal{M})$) denotes the set of vector (resp. covector) fields on $\mathcal{M}$.
Let $X,Y \in \Gamma(T\mathcal{M})$, the vector field $\nabla_Y X$ is the covariant derivative of $X$ with respect to $Y$.
In the context of dynamical systems subjected to external driving forces on manifolds, a force at a point $p \in \mathcal{M}$ is a covector, namely $f_d : T_p\mathcal{M} \times I \rightarrow T^*_p\mathcal{M}$.

The metric can be used to uniquely relate elements of $T\mathcal{M}$ and elements of $T_*\mathcal{M}$. For each $p \in \mathcal{M}$ we define the flat map $ (\cdot)^{\flat} : T_p\mathcal{M} \rightarrow T^*_p\mathcal{M}$ and sharp map $(\cdot)^{\sharp} : T^*_p\mathcal{M} \rightarrow T_p\mathcal{M}$ as the inverse of $(\cdot)^{\flat}$.

$C^{\infty}(\mathcal{M})$ denotes the set of smooth functions $\varphi : \mathcal{M} \rightarrow \mathbb{R}$.
The differential of a function $\varphi \in C^{\infty}(M)$ is the covector field $d \varphi \in \Gamma(T^*M)$.
In local coordinates, $d \varphi = \frac{\partial \varphi}{\partial x^i} dx^i$.
To express the partial derivative, we will adopt the contracted notation $\partial_i \varphi = \frac{\partial \varphi}{\partial x^i}$.

Let $\mathcal{M}$ and $\mathcal{N}$ be two differentiable Riemannian manifolds and $f: \mathcal{M} \rightarrow \mathcal{N}$ be a continuous map.
The \emph{pushforward} map $f_*$ is the map $f_* : T\mathcal{M} \rightarrow T\mathcal{N}$ where $f_*(X) \varphi := X(f \circ \varphi) \quad \forall \varphi \in C^{\infty}(\mathcal{N}), \forall X \in \Gamma(T\mathcal{M})$.
The \emph{pullback} map $f^*$ is the map $f^* : T^*\mathcal{N} \rightarrow T^*\mathcal{M}$, where $f^*(\omega) (X) := \omega(f_*(X)) \quad \forall X \in \Gamma(T\mathcal{M}), \forall \omega \in T^*\mathcal{N}$.

A curve $\gamma$ on a given manifold $\mathcal{M}$ is a mapping $\gamma: I \subset \mathbb{R} \rightarrow \mathcal{M}$.
The curve can be expressed in local coordinate through the mapping $x_{\gamma} = x \circ \gamma : I \rightarrow \mathbb{R}^d$ such that $x(\gamma(t)) = x_{\gamma}(t) \in \mathbb{R}^d$ for each $t \in I$.
We use $\dot{x}_{\gamma}(t)$ to express the \emph{speed} $\frac{d x_{\gamma}}{dt}$.
Wherever the explicit reference to the underlying curve is not needed, we use directly $\dot{x}^i$ to indicate the i-th local coordinate of the velocity of the curve.

Given a curve $\gamma : I \rightarrow \mathcal{M}$, a vector field along $\gamma$, $v_{\gamma}$, is a map that assigns to each $t \in I$ an element $v_{\gamma(t)} \in T_{\gamma(t)} \mathcal{M}$.
The covariant derivative of $v_{\gamma}$ along $v_{\gamma}$ in local coordinates~is
\begin{equation}
    \left(\nabla_{v_{\gamma}} v_{\gamma} \right)^k = \ddot{x}^k + \Gamma_{ij}^k \dot{x}^i \dot{x}^j,
    \label{eqn:cov_derivative}
\end{equation}
with $\Gamma_{ij}^k$ the Christoffel symbols.
A Riemannian metric $g$ induces a unique affine connection $\nabla$ on $\mathcal{M}$, called the Levi-Civita connection.
In this scenario the Christoffel symbols can be expressed as a function of the Riemannian metric $g$.
In local coordinates the Christoffel symbols for the Levi-Civita connection are $\Gamma_{ij}^k = \frac{1}{2} g^{km} \left( \partial_i g_{mj} + \partial_j g_{mi} - \partial_m g_{ij} \right)$ for $(i, j, k \in {1,\dots, d})$.

A second-order linear DS on $\mathcal{M}$ can be expressed, for $t \in I \subset \mathbb{R}$, in intrinsic formulation as
\begin{equation}
    \nabla_{v_{\gamma}} v_{\gamma} = \mathcal{F}(\gamma, v_{\gamma},t)^\# = -d\phi^\# - D(\cdot,v_{\gamma})^\#,
    \label{eqn:ds_intrinsic}
\end{equation}
where $\gamma : I \rightarrow \mathcal{M}$ is a curve on the manifold $\mathcal{M}$, $v_{\gamma}$ is the vector field generated by the tangent velocities of curve $\gamma$.
On the right-hand side of~\Cref{eqn:ds_intrinsic}, $\mathcal{F}$ is the total forces covector.
It can be further split into elastic and dissipative components.
Given a potential function, $\phi \in C^{\infty}(\mathcal{M})$, $d\phi^\#$ represents the elastic gradient field, while $D(\cdot,v_{\gamma}) \in T^*\mathcal{M}$ is the dissipative covector field.
In local coordinates we have
\begin{equation}
    \left(\nabla_{v_{\gamma}} v_{\gamma} \right)^k  = -g^{ak} \partial_a\phi - D_m^k \dot{x}^m.
    \label{eqn:second_ds}
\end{equation}
Combining~\Cref{eqn:cov_derivative,eqn:second_ds}, we have
\begin{equation}
    \overbrace{\ddot{x}^k}^{\ddot{\mathbf{x}}} + \overbrace{\Gamma_{ij}^k \dot{x}^i}^{\boldsymbol{\Xi}} \overbrace{\dot{x}^j}^{\dot{\mathbf{x}}}  = -\overbrace{g^{ak}}^{\m{G}^{-1}} \overbrace{\partial_a\phi}^{\nabla \phi} - \overbrace{D_m^k}^{\m D} \overbrace{\dot{x}^m}^{\dot{\mathbf{x}}}.
    \label{eqn:full_ds}
\end{equation}
Note that $D^{k}_m = g^{ak}D_{am}$.
\Cref{eqn:full_ds} can be expressed using vector notation as
\begin{equation}
    \ddot{\mathbf{x}} = \mathbf{f}(\mathbf{x},\dot{\mathbf{x}}) = -\m{G}^{-1} \nabla \phi -\m{D} \dot{\mathbf{x}} - \boldsymbol{\Xi} \td{\v{x}}.
    \label{eqn:vector_ds}
\end{equation}
In the following sections, we will use capital bold letters to represent matrices in vector notation.
When indices are used alongside a matrix, for the sake of clarity, the matrix itself will be denoted using capital letters without bold formatting.

\section{Learning the Latent Manifold Embedding}
\label{sec:learning_embedding}
Let $\mathcal{M}$ and $\mathcal{N}$ be two (non-compact) Riemannian manifolds, with respective charts $\rbr{\mathcal{U},x}$ and $\rbr{\mathcal{V},y}$.
We will indicate with $g$ and $h$, the Riemannian metrics of $\mathcal{M}$ and $\mathcal{N}$, respectively.
In particular, for $\text{dim}(\mathcal{M}) = d$, $\mathcal{N}$ coincides with a $(d+1)$-dimensional Euclidean space, $\mathbb{R}^{d+1}$.
Therefore, $\mathcal{V}\equiv \mathbb{R}^{d+1}$ and $y \equiv id_{\mathbb{R}^{d+1}}$, where $id_{\mathbb{R}^{d+1}}$ is the identity map.
Let $h = \delta_{ij}$ with respect to the chosen chart, where $\delta_{ij}$ is the Kronecker symbol, $\delta_{ij}=1$ if $i=j$ otherwise $\delta_{ij}=0$, and $i,j=1,\dots,d+1$.
\begin{figure}[ht]
    \centering
    \scalebox{.8}{\begin{tikzpicture}
    \node at (2,0) (manifold_m) {\Large $\mathcal{M}$};
    \node at (6,0) (manifold_n) {\Large $\mathcal{N} \equiv \mathbb{R}^{d+1}$};
    \node at (6,-2) (chart_n) {\Large$ \mathbb{R}^{d+1}$};
    \node at (2,-2) (chart_m) {\Large $\mathbb{R}^d$};

    \draw[-stealth] (manifold_m) -- node[above] {$f$} (manifold_n);
    \draw[-stealth] (manifold_n) -- node[right] {$(\mathcal{V}\equiv \mathbb{R}^{d+1},y \equiv id_{\mathbb{R}^{d+1}})$} (chart_n);
    \draw[-stealth] (manifold_m) -- node[right] {$(\mathcal{U},x)$} (chart_m);
    \draw[-stealth] (chart_m) -- node[above] {$y \circ f \circ x^{-1}$} (chart_n);
\end{tikzpicture}}
    \caption{\footnotesize Mapping structure across manifolds.}
    \label{fig:manifold_embedding}
\end{figure}
With reference to~\Cref{fig:manifold_embedding}, $f : \mathcal{M} \hookrightarrow \mathbb{R}^{d+1}$ is a smooth isometric embedding into a $d+1$ Euclidean space.
$(y \circ f \circ x^{-1})$ is the local coordinates expression of the embedding, i.e. the mapping between the two charts $(\mathcal{U},x)$ and $(\mathcal{V},y)$.

We propose to model the components of the local coordinates formulation of the embedding as follows
\begin{equation}
    (y \circ f \circ x^{-1})^i = \begin{cases}
        x^i \quad \text{if } i \le \text{dim}(\mathcal{M}) \\
        \psi(x^i;\mathbf{w}) \quad \text{otherwise}
    \end{cases},
    \label{eqn:embedding}
\end{equation}
where $\psi : \mathbb{R}^d \rightarrow \mathbb{R}$ is a smooth function parametrized by the weights $\v{w}$ and $\psi \in C^r(\mathbb{R}^d)$ with $r \ge 1$.
We will refer to the embedding expressed in local coordinates with the vector notation $\mathbf{\Psi}: \mathbb{R}^d \rightarrow \mathbb{R}^{d+1}$, highlighting the implicit dependence on $\psi$.
\begin{proposition}
    \label{prop:embedding}
    $f : \mathcal{M} \rightarrow \mathbb{R}^{d+1}$ is a smooth mapping with local coordinates as in~\Cref{eqn:embedding}. $f : \mathcal{M} \hookrightarrow \mathbb{R}^{d+1}$ is an embedding.
\end{proposition}
\begin{proof}
    \footnotesize
    See~\Cref{app:proof_prop}.
\end{proof}
All the geometric operators, namely the metric and the Christoffel symbols, in~\Cref{eqn:vector_ds}, are derived from the embedding defined in~\Cref{eqn:embedding} by leveraging on the pullback operation.
In order to improve readability, in the following, we drop the explicit dependency of $\psi$ on the local coordinates point $\v{x}$ and the weights $\v{w}$.
Nevertheless, all operators derived have to be considered dependent on $\v{x}$ and $\v{w}$ via $\psi$.

We first derive the metric as a function of the approximator $\psi$.
The components of the pushforward map, $J_j^i = \partial_j(y \circ f \circ x^{-1})^i$, can be expressed in matrix notation as
\begin{equation}
    \m{J}(\mathbf{x};\mathbf{w}) =
    \begin{bmatrix}
        \mathbf{I}^{\text{dim}(\mathcal{M}) \times \text{dim}(\mathcal{M})} \\
        \nabla \psi ^T
    \end{bmatrix},
    \label{eqn:jacobian}
\end{equation}
where $\nabla \psi = \left[ \partial_1 \psi, \dots, \partial_{\text{dim}(\mathcal{M})} \psi \right]^T$; $\m J$ is also known as the Jacobian.
Isometry of the embedding implies that the metric on $\mathcal{M}$ can be derived from the metric on  $\mathcal{N}$ via the pullback operation of the metric, $g = f^*h$.
In local coordinates this can be expressed as
\begin{equation}
    g_{ij} = \partial_i(y \circ f \circ x^{-1})^a h_{ab} \partial_j(y \circ f \circ x^{-1})^b.
    \label{eqn:pullback_metric}
\end{equation}
Given $h_{ab} = \delta_{ab}$, \Cref{eqn:pullback_metric} can be written in matrix notation as
\begin{equation}
    \m{G}(\mathbf{x};\mathbf{w}) = \m{J}^T\m{J} = \m{I} + \nabla \psi \nabla \psi^T.
    \label{eqn:pullback_metric_vec}
\end{equation}

To derive the Christoffel symbols, we need to express the derivative of the metric.
Given that the metric depends on $\mathbf{x}$ only through the term $\nabla \psi \nabla \psi^T$, the Christoffel symbols (contracted with the velocity) can be expressed as
\begin{align}
    \Xi^q_j(\mathbf{x}, \dot{\mathbf{x}};\mathbf{w}) = 
    & g^{qm} \frac{1}{2} \biggl( \partial_i \left( \partial_m \psi \pd{\psi}{j} \right) + \notag \\
    & \partial_j \left( \partial_m \psi \partial_i \psi \right) - \partial_m \left( \partial_i \psi \partial_j \psi \right) \biggr) \dot{x}^i
    \label{eqn:christoffel_second_kind}
\end{align}

The last two terms not yet defined in~\Cref{eqn:vector_ds} are the potential energy $\phi$ and the dissipative coefficients $\m{D}$.
We impose to the potential energy a classical quadratic structure $\phi = \frac{1}{2} \mathbf{x}^T \m{K} \mathbf{x}$.
Both the matrices $\m{K}$ and $\m{D}$ can be user-defined or left "open" for optimization.

Given the full state, $(\v{x},\v{\dot{x}})$, of a DS, our training dataset is composed of position-velocity pairs $\lbrace \mathbf{x}_m,\mathbf{\dot{x}}_m | m = 1,\dots,M \rbrace$, with $M$ the total number of sampled points.
The ground truth is given by sampled acceleration $\lbrace \mathbf{\ddot{x}}_m | m = 1,\dots,M \rbrace$.
We propose the following optimization problem
\begin{equation}
    \min_{\mathbf{w}, \m{K}, \m{D}} \mathcal{F}_2(\v{x}_m, \td{\v{x}}_m, \tdd{\v{x}}_m \vert \v w, \m K, \m D),
    \label{eqn:optim_second}
\end{equation}
with
\begin{align}
    \mathcal{F}_2(\v{x}_m, \td{\v{x}}_m, & \tdd{\v{x}}_m \vert \v w, \m K, \m D) = \notag                                                                                                  \\
    \sum_{m=1}^M \lVert           & \ddot{\mathbf{x}}_m + \m{G}^{-1}(\mathbf{x}_m;\mathbf{w}) \left( \m{K}(\mathbf{x}_m - \mathbf{x}^*) + \m{D} \dot{\mathbf{x}}_m \right) + \notag \\
                                         & \boldsymbol{\Xi} (\mathbf{x}_m, \dot{\mathbf{x}}_m;\mathbf{w}) \dot{\mathbf{x}}_m \rVert ^2 + \lambda \norm{\v w}^2,
    \label{eqn:optim_fun_second}
\end{align}
where $\m{K}, \m{D} \in \mathcal{S}^d_{++}$, the manifold of Symmetric Positive Definite (SPD) matrices of dimension $d$.
$\mathbf{x}^*$ is the fixed stable equilibrium point, or attractor, of the DS.
$\lambda$ is a parameter weighing the regularization term.
This parameter affects the manifold's curvature smoothness.
Since the manifold's curvature translates into acceleration within the DS (via the Christoffel symbols), regularization plays a crucial role in containing high frequency change of curvature, avoiding high accelerations and potential overfitting.

SPD matrices optimization is achieved by parametrazing  the generic SPD matrix as
\begin{equation}
    \m M(\boldsymbol{\alpha}, \boldsymbol{\xi}) = U(\boldsymbol{\alpha}) \Lambda(\boldsymbol{\xi}) U(\boldsymbol{\alpha})^T,
\end{equation}
with
\begin{equation}
    \Lambda(\boldsymbol{\xi}) = \begin{bmatrix}
        e^{\xi_1} & 0      & 0         \\
        0         & \ddots & 0         \\
        0         & 0      & e^{\xi_d}
    \end{bmatrix}
\end{equation}
and $\m U$ the orthogonal matrix resulting from QR decomposition of the matrix constructed so to have the first column equal to the vector $\boldsymbol{\alpha} \in \mathbb{R}^d$. $\boldsymbol{\alpha}$ and $\boldsymbol{\xi}$ are the learnable parameters.

\begin{theorem}
    \label{thm:stability}
    Let $\mathbf{w} \in \rbr{-\infty, \infty}$. The dynamical system
    \begin{equation}
        \ddot{\mathbf{x}} = -\m{G}^{-1}(\mathbf{x};\mathbf{w}) \left(\m{K}(\mathbf{x} - \mathbf{x}*) + \m{D} \dot{\mathbf{x}} \right) - \boldsymbol{\Xi} (\mathbf{x},\dot{\mathbf{x}};\mathbf{w}) \dot{\mathbf{x}}
    \end{equation}
    is globally asymptotically stable at the attractor $\mathbf{x}*$, i.e. $\lim_{t \rightarrow \infty} \norm{\mathbf{x} - \mathbf{x}*} = 0$.
\end{theorem}
\begin{proof}
    \footnotesize
    See~\Cref{app:proof_thm}.
\end{proof}

\subsection{Gradient Systems \& Incremental Learning}
\label{subsec:gradient_systems}
Our method can be applied to first order dynamics.
Let $X : \mathcal{M} \rightarrow T\mathcal{M}$ be a vector field on $\mathcal{M}$. $X$ is called a gradient system if
\begin{equation}
    X = -d\phi^{\#} \quad \overset{\text{local coordinates}}{\longrightarrow} \quad X^k = -g^{ik} \partial_i\phi.
    \label{eqn:first_ds}
\end{equation}
As for the system in~\Cref{eqn:second_ds}, gradient systems are globally stable; moreover, they are globally exponentially stable.
Such property follows from the fact that this type of systems are contracting on $\mathcal{M}$, \cite{simpson-porco_contraction_2014}, i.e. $g(\nabla_v X, v) \le -\lambda g(v,v)$ for $p \in \mathcal{U}$ and $v \in T_p\mathcal{M}$. $\lambda > 0$ is the contraction rate.
For strongly convex functions $\phi$ on $\mathcal{U}$, they satisfy the contraction condition $\text{Hess}(\phi) \succeq \lambda g$, where $\text{Hess}(\phi)$ is the Riemannian Hessian, see \cite{simpson-porco_contraction_2014,wensing2020beyond} for details.

We can minimize the Mean Square Error (MSE) loss, as in~\Cref{eqn:optim_second}, having as target the sampled velocities
\begin{equation}
    \min_{\mathbf{w}, \m{K}} \sum_{m=1}^M \mathcal{F}_1(\v{x}_m, \td{\v{x}}_m \vert \v w, \m K).
    \label{eqn:optim_first}
\end{equation}
with
\begin{align}
     & \mathcal{F}_1(\v{x}_m, \td{\v{x}}_m \vert \v w, \m K) = \notag                                                                            \\
     & \sum_{m=1}^M \norm{\dot{\mathbf{x}}_m + \m{G}^{-1}(\mathbf{x}_m;\mathbf{w}) \m{K}(\mathbf{x}_m - \mathbf{x}*)}^2 + \lambda \norm{\v w}^2,
    \label{eqn:optim_fun_first}
\end{align}

If for simpler scenarios, first-order DS might suffice, more complicated cases require the usage of second-order DS.
Nevertheless, first-order DS optimization can be performed as a way of finding a good initial solution for the second-order DS.
As shown in~\Cref{sec:synthetic_eval}, given the simple structure of a first-order DS, solution to~\Cref{eqn:optim_first} can be found in considerably lower time than~\Cref{eqn:optim_second}.
For complicated problems, this situation suggests to perform a sort of incremental learning: first, find optimal embedding weights, $\mathbf{w}*$ and stiffness matrix, $\mathbf{K}*$, by solving repeatedly~\Cref{eqn:optim_first}; second, solve~\Cref{eqn:optim_second} starting from $\mathbf{w}*$ and $\mathbf{K}*$.

\section{Online Kernel-Based Local Space Deformation}
\label{sec:kernel_deformation}
Kernel-based space deformation is an effective method for generating localized curvature, which subsequently influences the behavior of chart-based DS.
As detailed in~\Cref{sec:obstacle_avoidance}, for obstacle avoidance scenarios, this approach is particularly relevant to our method, where an explicit representation of the latent manifold is available.

Nevertheless, kernel-based deformation can also be adopted to model any type of global manifold's curvature realized as a linear combination of point-wise sources of the deformation.
The analytical formulation provides a simplified framework that would allow us to gain intuition on the effect of the metric tensor and Christoffel symbols in the chart space DS, as curvature starts to appear in the manifold.
These concepts are explored in~\Cref{sec:first_deformation,sec:second_deformation}.
Starting from a flat space scenario, we analyze the effect of locally deforming the space via Radial Basis Function (RBF) kernels for first-order dynamics and second-order dynamics.
In addition, for second-order dynamics, we show how it is possible to leverage on hybrid harmonic and geodesic motion to achieve concave obstacle avoidance, in fairly extreme scenarios, without recurring to planning strategies.

\subsection{Obstacle Avoidance via Direct Space Deformation}
\label{sec:obstacle_avoidance}
One advantage of the proposed method is the direct extendibility to obstacle avoidance scenarios.
We take advantage of the geometric obstacle avoidance techniques based on the local deformation or stretching of the space.
In our approach, we encode the non-linearity of the DS within the curvature of the space.
Without repeating the learning process, the non-linearity needed for the obstacle avoidance task can be encoded, locally, in the curvature of the space as well.

Geometry-based obstacle avoidance techniques rely on the definition of a metric that takes into account the presence of obstacles.
The metric can be defined in the ambient space (and pulled back afterwards), \cite{beik-mohammadi_learning_2021}, or directly in chart space, \cite{cheng_rmpflow_2020}. The two approaches are equivalent.
They work well if the original space on which the DS is taking place does not present pre-existent relevant curvature.
In our approach, an explicit knowledge of the "shape" of the manifold is available.
Given this knowledge, we show how it is possible to directly deform an already non-linear space to produce obstacle avoidance.

Let $k_{\text{chart}}: \mathbb{R}^d \times \mathbb{R}^d \rightarrow \mathbb{R}$ ($k_{\text{ambient}}: \mathbb{R}^{d+1} \times \mathbb{R}^{d+1} \rightarrow \mathbb{R}$) be a similarity measure in the chart (ambient) space.
Let $\bar{\mathbf{x}} \in \mathbb{R}^d$ ($\bar{\mathbf{y}} \in \mathbb{R}^{d+1}$) be the location of an obstacle in chart (ambient) space.
Given a current position $\mathbf{x} \in \mathbb{R}^d$ ($\mathbf{y} \in \mathbb{R}^{d+1}$), $k(\mathbf{x}, \bar{\mathbf{x}})$ ($k(\mathbf{y}, \bar{\mathbf{y}})$) informally expresses how close we are to the obstacle in the chart (ambient) space.
Considering $\bar{\mathbf{x}}$ ($\bar{\mathbf{y}}$) fixed, we have $k_{\text{chart}}(\bar{\mathbf{x}}, \cdot): \mathbb{R}^d \rightarrow \mathbb{R}$ ($k_{\text{ambient}}(\bar{\mathbf{y}}, \cdot): \mathbb{R}^{d+1} \rightarrow \mathbb{R}$).
We assume $k_{\text{chart}} \in C^r(\mathbb{R}^d)$ ($k_{\text{ambient}} \in C^r(\mathbb{R}^{d+1})$) with $r \ge 1$ and $k_{\text{chart}}(\mathbf{x}, \bar{\mathbf{x}}) \approx 0$ ($k_{\text{ambient}}(\mathbf{y}, \bar{\mathbf{y}}) \approx 0$) for $\norm{\mathbf{x} - \bar{\mathbf{x}}} \ge \epsilon$ ($\norm{\mathbf{y} - \bar{\mathbf{y}}} \ge \epsilon$) for some $\epsilon > 0$.
The first condition imposes at least one time differentiability while the second one requires fast decay of the similarity measure away from the obstacle.

Recall in~\Cref{eqn:pullback_metric,eqn:pullback_metric_vec}, we assumed the ambient metric to be constant over all the space and equal to the identity, namely the Euclidean metric.
We now use the following metric for the ambient Euclidean space
\begin{equation}
    \m{H} = \m{I} + \nabla k_{\text{ambient}} \nabla k_{\text{ambient}},
    \label{eqn:ambient_metric}
\end{equation}
where $\nabla k_{\text{ambient}}$ expresses the derivative of the similarity measure $k_{\text{ambient}}$ with respect to $\mathbf{y}$.

The metric in~\Cref{eqn:ambient_metric} is implicitly defining a deformation of the ambient space.
It can be derived via the pullback of the Euclidean metric of a $d$+$2$-dimensional Euclidean space, embedding our ambient space $\mathbf{\Psi}_{\text{ambient}} = [\mathbf{y}, k(\bar{\mathbf{y}},\mathbf{y})]^T$.
With $\m{H} \in \mathbb{R}^{\text{dim}(\mathcal{M}) + 1 \times \text{dim}(\mathcal{M}) + 1}$, the chart space metric can be derived via the pullback of the ambient metric as in~\Cref{eqn:pullback_metric}, $\m{G}(\mathbf{x};\mathbf{w}) = \m{J}^T\m{H}\m{J}$.

\begin{figure*}[t]
    \centering
    \subfloat[]{\includegraphics[width=.22\textwidth]{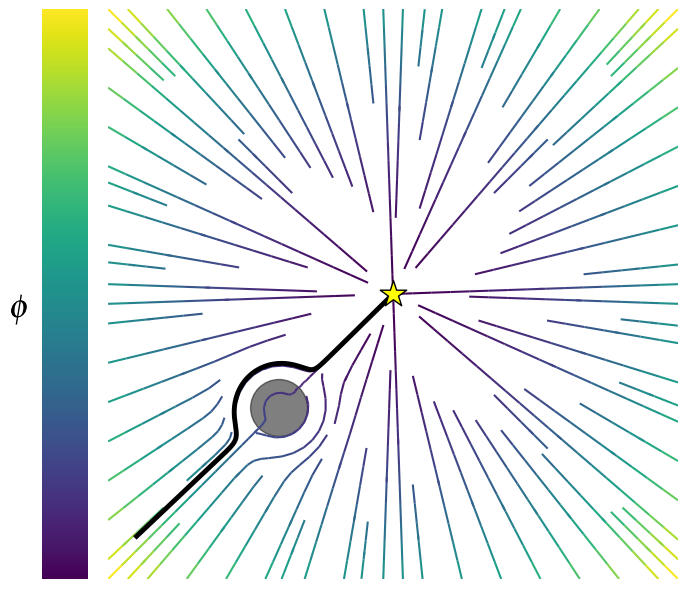}
        \label{fig:convex_field_first}}
    \hfill
    \subfloat[]{\includegraphics[width=.21\textwidth]{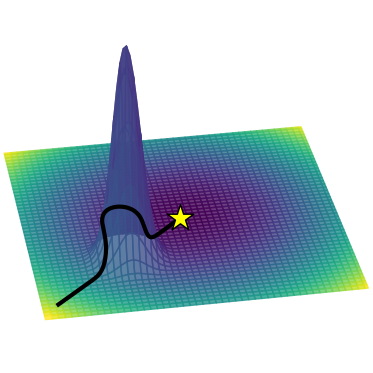}
        \label{fig:convex_embedding_first}}
    \hfill
    \subfloat[]{\includegraphics[width=.19\textwidth]{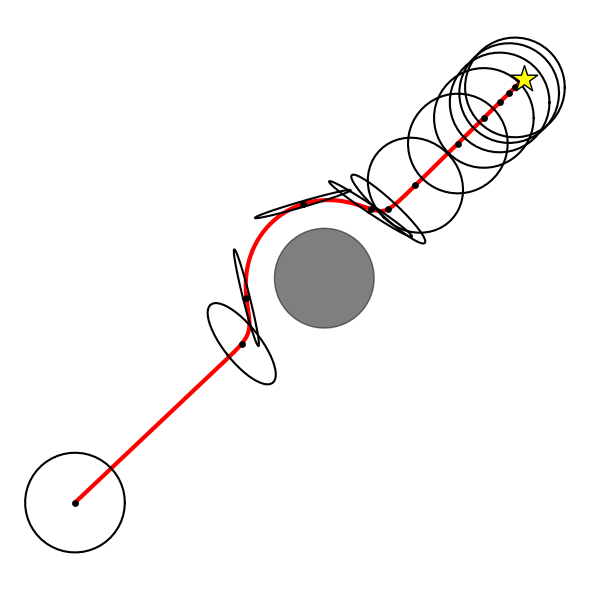}
        \label{fig:convex_metric_first}}
    \hfill
    \subfloat[]{\includegraphics[width=.23\textwidth]{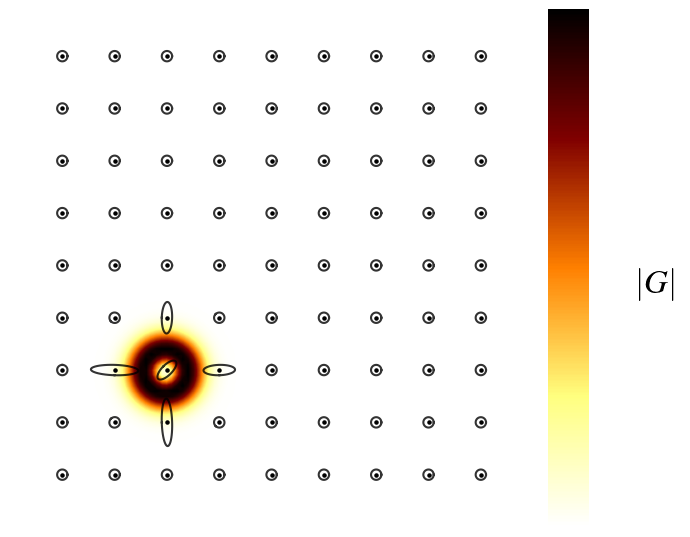}
        \label{fig:convex_detmetric_first}}
    \caption{\footnotesize Firs-order DS in flat space with localized deformation in the obstacle area: (a) Vector field with one sampled streamline avoiding the obstacle; (b) $3$D embedded representation of the manifold; (c) one sampled trajectory with eigenvalue decomposition ellipses of the inverse of the metric for selected location; (d) metric determinant function with eigenvalue decomposition ellipses of the metric.}
    \label{fig:convex_first}
\end{figure*}

The pullback operation is equivalent to adding the "obstacle" metric to the chart space metric encoding the non-linearity of the DS, $\m{G}_{\text{total}} = \m{G}(\mathbf{x};\mathbf{w}) + \m{G}_{\text{obs}}$, where $\m{G}_{\text{obs}} = \m{I} + \nabla k_{\text{chart}} \nabla k_{\text{chart}}$. Let $\m{H} = \begin{bmatrix}
        \m{I} + \m{A} & \m{B}     \\
        \m{B}^T       & 1 + \m{C}
    \end{bmatrix}$,
with $\m{A} \equiv \nabla k_{\text{chart}} \nabla k_{\text{chart}} \in \mathbb{R}^{d \times d}$, $\m{B} \in \mathbb{R}^{1 \times d}$ and $\m{C} \in \mathbb{R}^{1 \times 1}$.
The pullback operation in~\Cref{eqn:ambient_metric} becomes $\m{G} = \m{I} + \m{A} + \m{B} \nabla \psi^T + \nabla \psi \m{B}^T + \nabla \psi (\m{I}+\m{C}) \nabla \psi^T$.
For $\gamma(t) \in \mathcal{M} \quad \forall t \ge 0$ we have $\m{B},\m{C} = 0$; therefore $\m{G} = \m{I} + \m{A} + \nabla \psi \nabla \psi^T = \m{I} + \nabla k_{\text{chart}} \nabla k_{\text{chart}}^T + \nabla \psi \nabla \psi^T = \m{G}_{\text{obs}} + \m{G}(\mathbf{x};\mathbf{w})$.

Adding metrics linearly is akin to treating the deformations of space, due to the non-linearity of the DS and the presence of an obstacle, separately.
$\m{G}(\mathbf{x};\mathbf{w})$ is derived as in~\Cref{eqn:pullback_metric_vec} from the embedding $\mathbf{\Psi}$ in~\Cref{eqn:embedding}; $\m{G}_{\text{obs}}$ can be derived from an embedding of the type $\mathbf{\Psi}_{\text{obstacle}} = [\mathbf{x}, k(\bar{\mathbf{x}},\mathbf{x})]^T$.
In other terms, the deformation of the space due to the presence of the obstacle is agnostic of the previous curvature in the manifold.
Such a scenario still yields good results where the space is fairly flat.

The explicit formulation of the embedding allows us to directly deform the space while actively taking into consideration the curvature of the space.
Let $\lbrace \bar{\mathbf{x}}_i, i = 1,\dots, N\rbrace$ be the location of $N$ obstacles in chart space; $\bar{\mathbf{w}}$ are the weights after learning the DS.
We model obstacles as a kernel-based local deformation of the spaces given by
\begin{equation}
    \bar{\psi}(\mathbf{x}) = \sum_{i=1}^N \eta_i k(\bar{\mathbf{x}}_i, \mathbf{x}),
    \label{eqn:local_deformation}
\end{equation}
where $\eta_i$ is a user-defined weight assigned to the local deformation at $\bar{\mathbf{x}}_i$.
At query-time the embedding in~\Cref{eqn:embedding} can be expanded as follows
\begin{equation}
    \mathbf{\Psi} = \begin{bmatrix}
        \mathbf{x} \\
        \psi(\mathbf{x};\bar{\mathbf{w}}) + \bar{\psi}(\mathbf{x}).
    \end{bmatrix}
    \label{eqn:embedding_obstacle}
\end{equation}
The embedding in~\Cref{eqn:embedding_obstacle} leads to the following pullback metric
\begin{equation}
    \m{G}(\mathbf{x};\mathbf{w}) = \mathbf{I} + \nabla \psi \nabla \psi^T + \overbrace{2 \nabla \psi \nabla \bar{\psi}(\mathbf{x})^T}^{\text{coupling term}} + \nabla \bar{\psi}(\mathbf{x}) \nabla \bar{\psi}(\mathbf{x})^T.
    \label{eqn:direct_deformation_metric}
\end{equation}
The coupling term in~\Cref{eqn:direct_deformation_metric} encodes the pre-existent curvature of the space.
Given the imposed condition on the regularity of $k$, both~\Cref{prop:embedding} and~\Cref{thm:stability} still hold.
The overall DS generated with such embedding retains global asymptotical stability, independently from the number of obstacles present.

\begin{figure*}[t!]
    \centering
    \subfloat[]{\includegraphics[width=.28\textwidth]{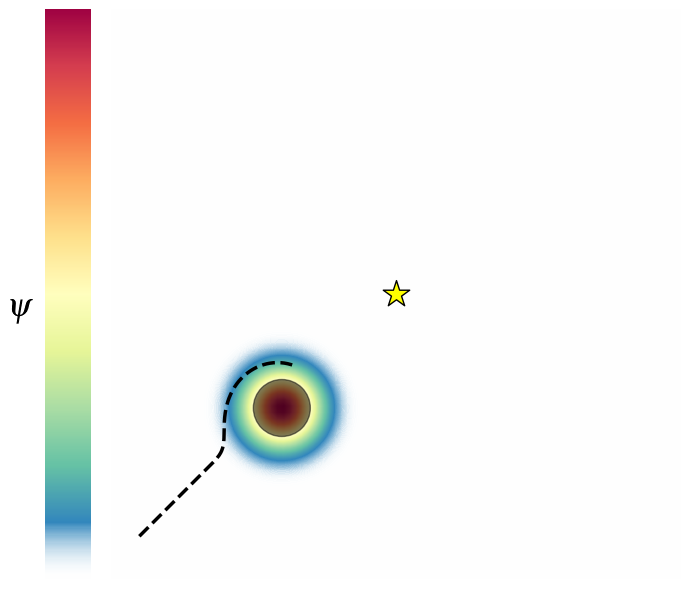}
        \label{fig:convex_field_second_geodesic_1}}
    \hfill
    \subfloat[]{\includegraphics[width=.24\textwidth]{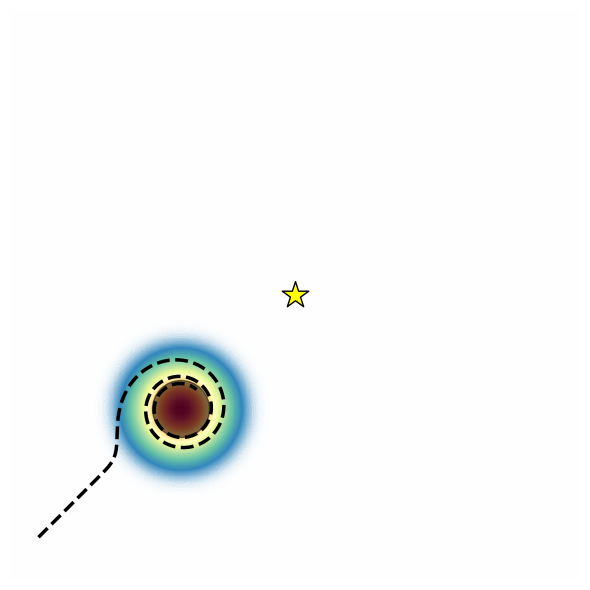}
        \label{fig:convex_field_second_geodesic_3}}
    \hfill
    \subfloat[]{\includegraphics[width=.24\textwidth]{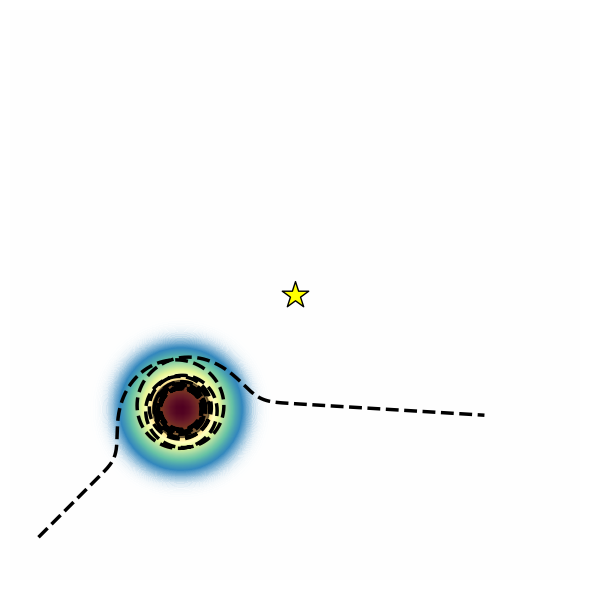}
        \label{fig:convex_field_second_geodesic_4}}
    \hfill
    \subfloat[]{\includegraphics[width=.21\textwidth]{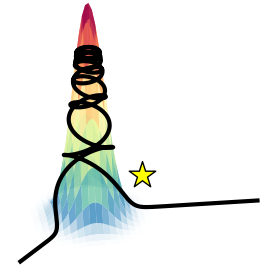}
        \label{fig:convex_field_second_embedding}}
    \caption{\footnotesize (a)-(c) Geodesic motion at time instants: $1$s, $5$s and $10$s; (d) 3D embedded representation of case (c). In background the color gradient represents the $d$+1 embedding coordinate.}
\end{figure*}

Second order dynamical systems cannot perform proximal obstacle avoidance.
Indeed, the geometrical term given by the Christoffel symbols generates forces that lead the system to "climb up" regions of local high-curvature, penetrating the obstacle.
This forces conflict with the potential and dissipative ones---$\m{G}^{-1} \nabla \phi$ and $\m{D} \dot{\mathbf{x}}$ in~\Cref{eqn:vector_ds}---generating an overall motion that stagnates right after the obstacle.
We refer the reader to~\Cref{sec:second_deformation} for more details and illustrations of these scenarios.
This issue can be solved by introducing velocity-dependent local space deformations
\begin{equation}
    \bar{\psi}(\mathbf{x}, \dot{\mathbf{x}}) = \sum_{i=1}^N \eta_i(\v x - \v{\bar{x}}_i,\dot{\mathbf{x}}) k(\bar{\mathbf{x}}_i, \mathbf{x}),
    \label{eqn:dynamic_deformation}
\end{equation}
where $\eta_i(\v x - \v{\bar{x}}_i,\dot{\mathbf{x}})$\footnote{Note that this function, though dependent on $\v x$, must be treated as constant in the derivation of the geometrical terms.} is framed as generalized sigmoid function acting on the cosine kernel between $\v x - \v{\bar{x}}_i$ and~$\dot{\mathbf{x}}$
\begin{equation}
    \eta_i(\v x - \v{\bar{x}}_i,\dot{\mathbf{x}}) = \frac{1}{1 + e^{-\tau \rbr{k_{\cos}(\v x - \v{\bar{x}}_i,\dot{\mathbf{x}}) - \cos{\theta_{ref}}}}},
    \label{eqn:sigmoid_fun}
\end{equation}
with $k_{\cos}(\v x - \v{\bar{x}}_i,\dot{\mathbf{x}}) = \frac{(\mathbf{x}-\mathbf{x}_i)^T\dot{\mathbf{x}}}{\norm{\mathbf{x}-\mathbf{x}_i}\norm{\dot{\mathbf{x}}}}$.
$\tau$ regulates the growth rate and $\theta_{ref}$ defines the starting growth point.
Safe option can be $\theta_{ref} = \frac{\pi}{2}$.
In this scenario, the underlying manifold dynamically deforms (locally) whenever the motion is monotonically decreasing towards the obstacle.
If the system is moving away from the obstacle, the manifold curvature goes back to the flat or nominal state.
This allows to effectively turn off the local geometrical terms once crossed the obstacle, allowing the system to reach the desired equilibrium point.

\subsection{Local Space Deformation In First Order DS}
\label{sec:first_deformation}
Let the $d+1$ embedding component $(y \circ f \circ x^{-1})^{d+1} = \psi$ be defined as a weighted sum of exponentially decaying kernels
\begin{equation}
    \psi(\mathbf{x}) = \sum_{i=1}^N \alpha_i k(\mathbf{x}, \bar{\mathbf{x}}_i) = \sum_{i=1}^N \alpha_i \exp{-\frac{\norm{\mathbf{x} - \bar{\mathbf{x}}_i}^2}{2 \sigma^2}},
\end{equation}
where $\bar{\mathbf{x}}_i$ is the $i$-th kernel center and $N$ is the number of kernel used. $\sigma$ and $\alpha_i$ are user-defined parameters; the former controls till which distant the local deformation affects the space geometry, the latter defines the magnitude of the deformation. In order to study the influence of a generic source of deformation, we consider $N=1$ and $\alpha_1=1$.

Via the pull-back of the embedding metric we recover the metric onto the manifold. In case of Euclidean (identity) metric for the ambient space we have
\begin{equation}
    \m{G}(\mathbf{x}) = \mathbf{I} + \frac{1}{\sigma^4}(\mathbf{x} - \bar{\mathbf{x}})(\mathbf{x} - \bar{\mathbf{x}})^T k(\mathbf{x}, \bar{\mathbf{x}})^2.
    \label{eqn:diff_source}
\end{equation}
See~\Cref{app:derivation_metric} for derivation details.
From~\Cref{eqn:diff_source} we note that the metric tensor is composed by two terms: an identity term, independent from the location of the deformation source, and a term active only in the neighborhood of the deformation where $k(\mathbf{x}, \bar{\mathbf{x}})^2 \approx 1$.
\begin{figure}[ht]
    \centering
    \resizebox{.2\textwidth}{!}{\begin{tikzpicture}
    \node[fill,circle,label={below:$\mathbf{x}$}, inner sep=0, minimum size=2mm] at (0,0) (x) {};
    \node[fill,circle,label={above:$\bar{\mathbf{x}}$}, inner sep=0, minimum size=2mm] at (1,1) (x_bar) {};
    \node at (1,-1) (x_ort) {};

    \draw[-stealth] (x) -- node[below,right] {$\lambda_1 = \norm{\v x - \bar{\v x}}^2$} (x_bar);
    \draw[-stealth] (x) -- node[right] {$\lambda_2 = 0$} (x_ort);
\end{tikzpicture}}
    \caption{\footnotesize Kernel-based metric tensor eigenvalues and eigenvectors.}
    \label{fig:eigen_metric}
\end{figure}
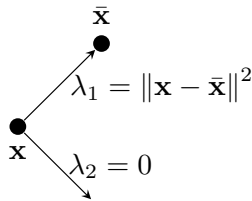
This second term is given by the outer product of the distance vector between the current location and the source of the deformation.
Outer product matrices are rank deficient with the eigenvector related to the only non-zero eigenvalues, $\lambda_1 = \norm{\v x - \bar{\v x}}^2$, directed as $\v x - \bar{\v x}$.

Consider a 2D space locally deformed in $\bar{\v x}$.
\Cref{fig:eigen_metric} shows the eigenvalues and the eigenvectors of the second term in the sum of~\Cref{eqn:diff_source}.
Recall that the metric tensor is used to measure lengths. In the direction of the deformation source, the space elongates of an amount proportional to $\norm{\v x - \bar{\v x}}^2 k(\mathbf{x}, \bar{\mathbf{x}})^2$.
In the direction perpendicular to the deformation source, as expected, the space does not elongate; indeed, $\lambda_2 = 0$. The space is only stretched in the direction of the deformation source.
This stretch reaches its maximum in the neighborhood of the source to than decrease gradually towards zero at $\bar{\mathbf{x}}$ where the space goes back to be flat, \Cref{fig:convex_detmetric_first}.
This explains clearly how obstacle avoidance is achieved for a gradient system as in~\Cref{eqn:first_ds}.
The projection of the gradient system's velocity onto the inverse of the metric tensor decreases the velocity's component in the direction of the obstacle, located at the source of deformation, of an amount inversely proportional to the entity of space stretching.
\Cref{fig:convex_metric_first} shows the ellipsoids generated by the inverse of the metric tensor.
The velocity component perpendicular to the source of the deformation remains unchanged.
This turns into an overall behavior of the gradient system that increases its velocity in the direction tangential to the obstacle.
Note that, if the velocity of the gradient system, $\v v$, points exactly towards the source of deformation, $\v v \parallel \v x - \bar{\v x}$, the streamline will not be deflected at all.
In such a case $\v v^T (\v x - \bar{\v x})^{\perp} = 0$.

\subsection{Local Space Deformation In Second Order DS}
\label{sec:second_deformation}
Second-order systems' behavior is affected by the Christoffel symbols.
This term depends on the derivative of the metric tensor.
As done for the differential of $\psi$, we can calculate the differential of the metric tensor
\[\label{eqn:second_diff_source}
    d\m{G}(\mathbf{x})[\mathbf{v}] &= \frac{1}{\sigma^4}\biggl( (\mathbf{v}\tilde{\mathbf{x}}^T + \tilde{\mathbf{x}}\mathbf{v}^T)k(\tilde{\mathbf{x}}) \notag \\
    &  + \frac{1}{\sigma^4}\tilde{\v x}\tilde{\v x}^T\rbr{\v{v}^T \tilde{\v x}\tilde{\v x}^T \v{v}} \biggr)k(\tilde{\mathbf{x}})^2.
\]
where $\tilde{\mathbf{x}} = \v x - \bar{\v x}$, see~\Cref{app:derivation_christoffel} for details.
Consider a geodesic motion.
The Christoffel symbol generates a deceleration perpendicular to the source of deformation.
\begin{figure}[ht]
    \centering
    \subfloat[]{\includegraphics[width=.225\textwidth]{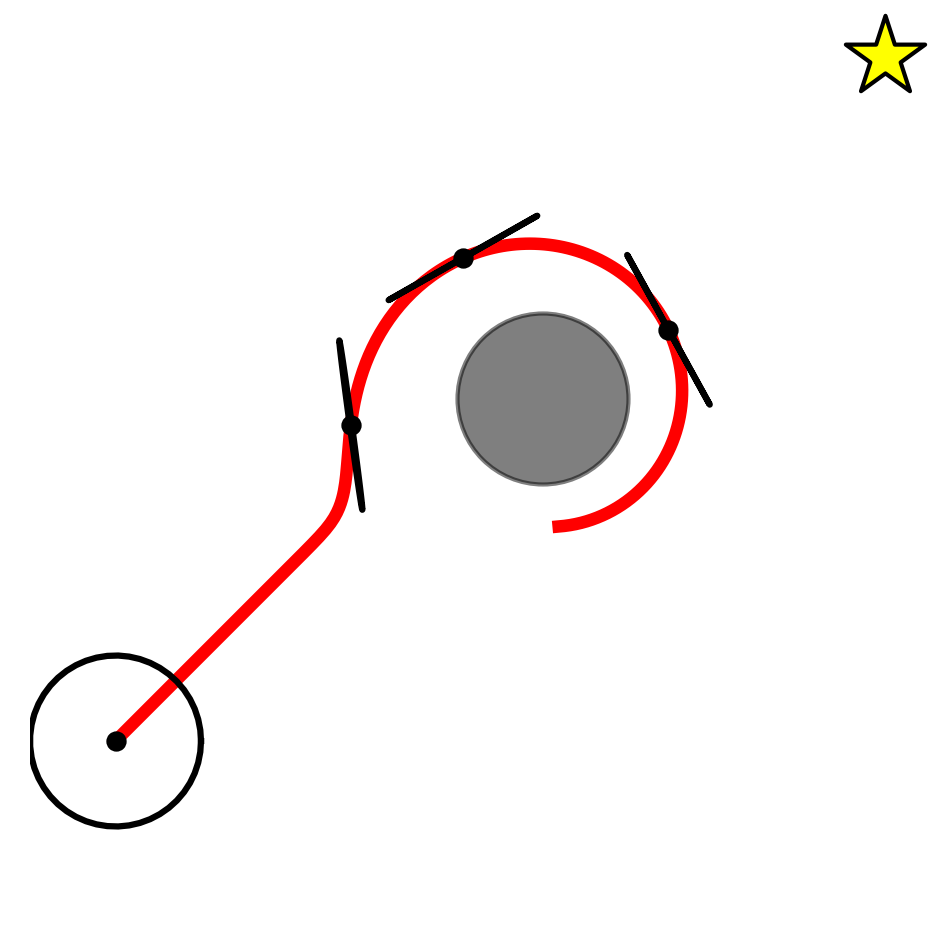}
        \label{fig:convex_metric_second_geodesic}}
    \hfill
    \subfloat[]{\includegraphics[width=.225\textwidth]{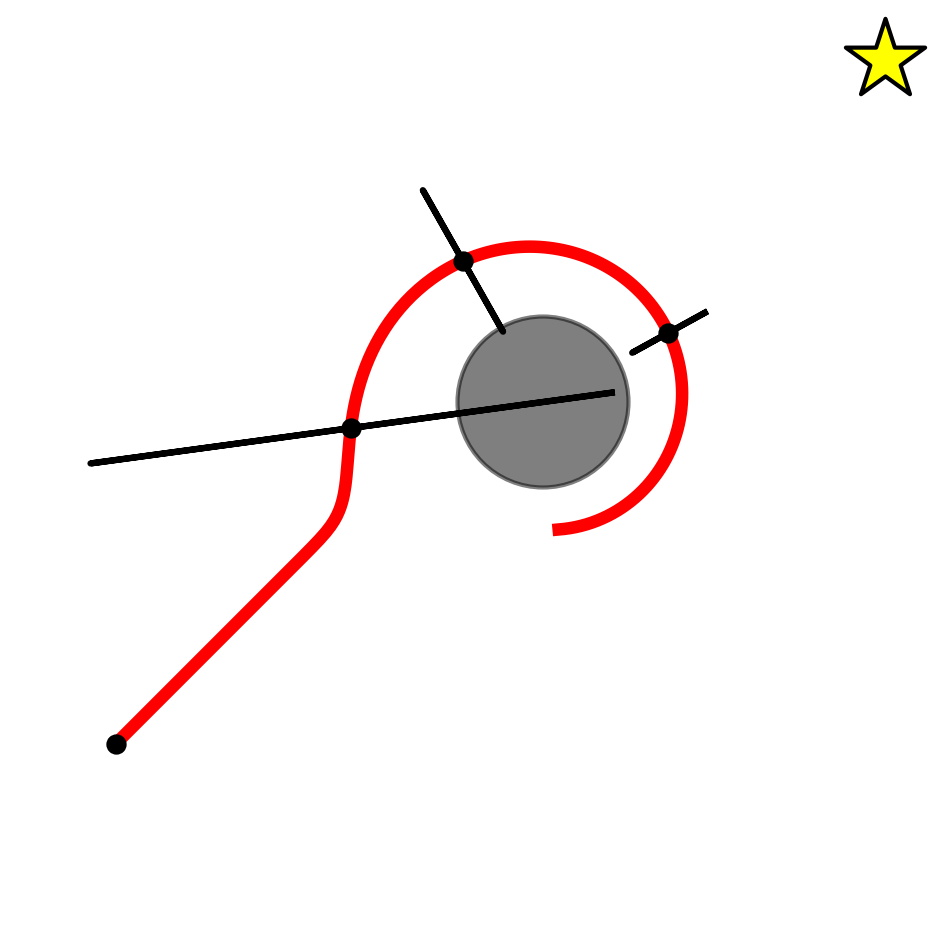}
        \label{fig:convex_christoffel_second_geodesic}}
    \caption{\footnotesize Geodesic motion after $1$s with ellipses representing eigenvalues and eigenvectors of (a) the inverse of the metric and (b) the Christoffel symbols for selected locations.}
\end{figure}

\begin{figure*}[t]
    \centering
    \subfloat[]{\includegraphics[width=.26\textwidth]{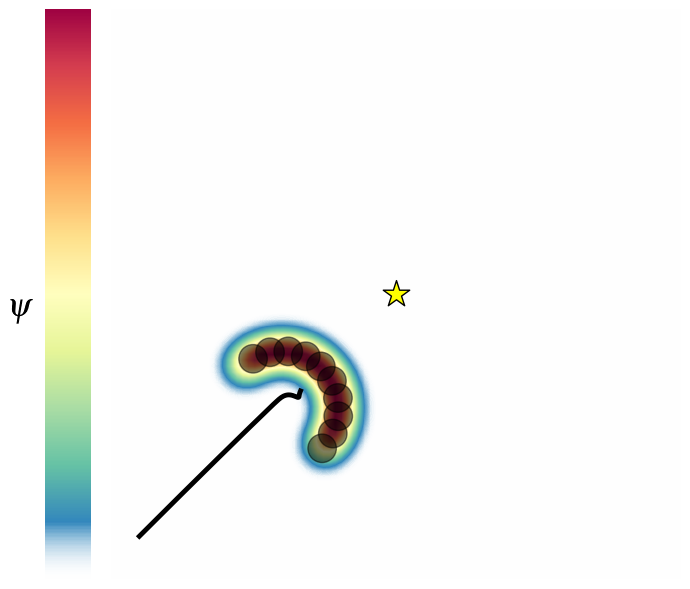}
        \label{fig:local_field_second_concave_harmonic}}
    \hfill
    \subfloat[]{\includegraphics[width=.22\textwidth]{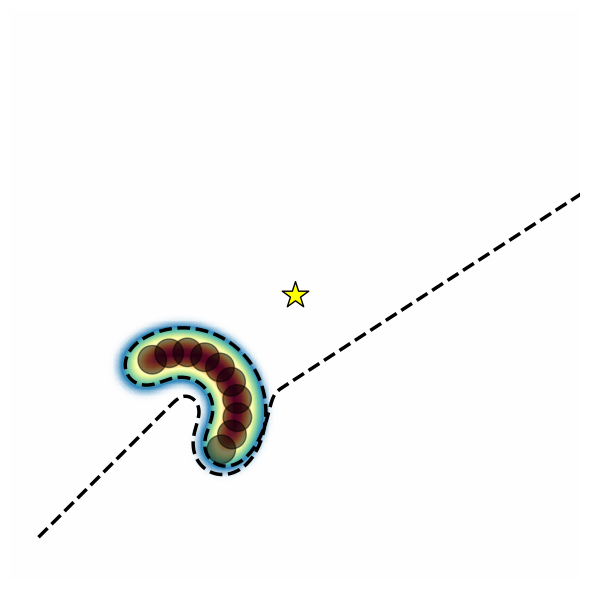}
        \label{fig:local_field_second_concave_geodesic}}
    \hfill
    \subfloat[]{\includegraphics[width=.22\textwidth]{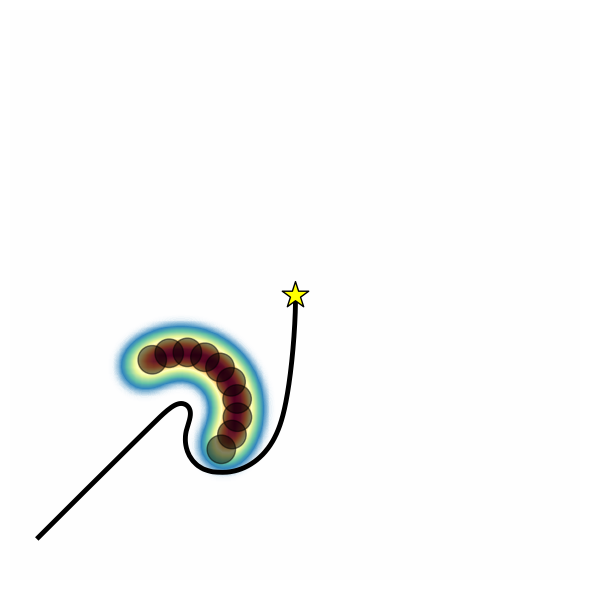}
        \label{fig:local_field_second_concave_hybrid}}
    \hfill
    \subfloat[]{\includegraphics[width=.22\textwidth]{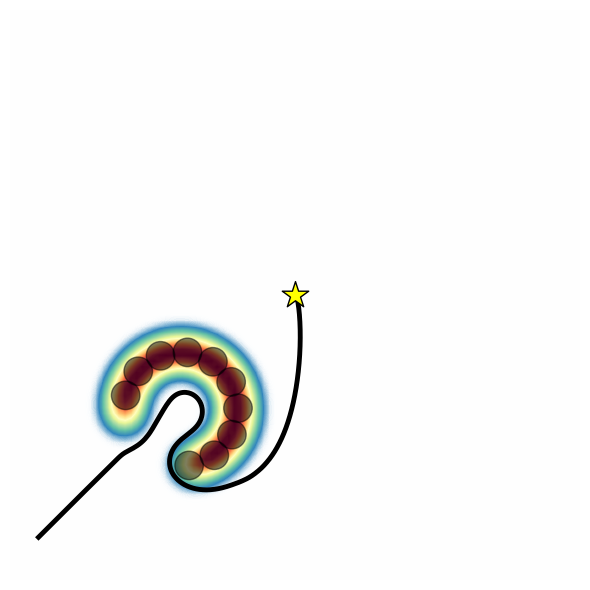}
        \label{fig:local_field_second_concave_hybrid_extreme}
    }
    \caption{\footnotesize Concave obstacle avoidance: (a) harmonic motion; (b) geodesic motion; (c)-(d) hybrid motion for semicircle and horseshoe obstacles.}
    \label{fig:dynamic_deformation_concave}
\end{figure*}

This, when approaching the source of deformation, deflects the streamlines avoiding the high curvature region.
Nevertheless, if the streamline transits too close to the source of deformation, the geodesic gets captured by the high curvature region.
\Cref{fig:convex_field_second_geodesic_1,fig:convex_field_second_geodesic_3,fig:convex_field_second_geodesic_4} show different frame of such geodesic motion.
This is clear by analyzing the Christoffel symbols' principal components shown in~\Cref{fig:convex_christoffel_second_geodesic}.
This components are perpendicular to the inverse metric ones, \Cref{fig:convex_metric_second_geodesic}, and they generate an inward acceleration towards the obstacle that "captures" the geodesic motion leading the streamline to climb up and down the source of deformation as illustrated in~\Cref{fig:convex_metric_second_geodesic}.

\begin{figure}[ht]
    \centering
    \subfloat[]{\includegraphics[width=.24\textwidth]{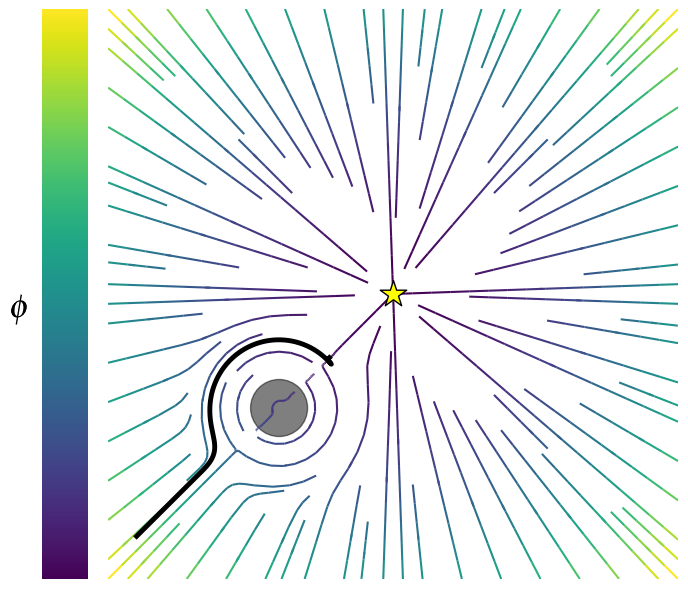}
        \label{fig:convex_field_second_harmonic_static}}
    \hfill
    \subfloat[]{\includegraphics[width=.21\textwidth]{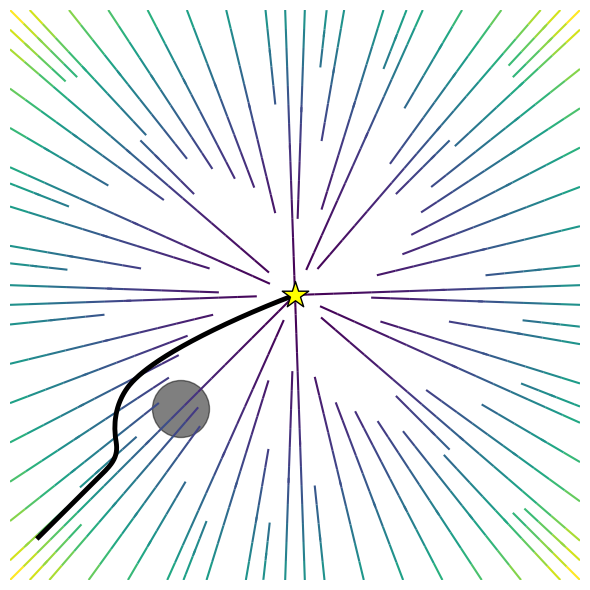}
        \label{fig:convex_field_second_harmonic_dynamic}}
    \caption{\footnotesize Second-order DS convex obstacle avoidance: (a) velocity independent local deformation; (b) velocity dependent deformation.}
    \label{fig:dynamic_deformation}
\end{figure}
When adopting second-order DS, the harmonic part conflicts with the Christoffel symbol, generating a stagnation of the DS right after the obstacle, \Cref{fig:convex_field_second_harmonic_static}.
In order to alleviate this issue, as seen in~\Cref{sec:obstacle_avoidance}, it is possible to define a velocity-dependent local deformation, \Cref{eqn:dynamic_deformation}.
This makes the curvature of the space "dynamic", so that it disappears when the obstacle is surpassed.
In this scenario, the effect of the Christoffel symbols, due to the local deformation, is canceled; the second-order DS behaves like a standard Euclidean harmonic oscillator, successfully reaching the attractor, \Cref{fig:convex_field_second_harmonic_dynamic}.
As a byproduct of this strategy, we obtained a more effective obstacle avoidance behavior.
The DS shows an asymmetrical behavior before and after the obstacle, following the most efficient trajectory to reach the attractor once surpassed the obstacle.

Concave obstacle avoidance represents a more challenging scenario where both first and second order DSs will stagnates in the middle of obstacle due to conflicting forces, \Cref{fig:local_field_second_concave_harmonic}.
Nevertheless, as seen previously, second-order system geodesics exhibit the ability of navigating through the space deformation.

Similarly to what seen before, \Cref{fig:local_field_second_concave_geodesic} shows the geodesic motion in face of concave obstacle.
When approaching the obstacle, the geodesic motion exhibits the ability of successfully avoid the deformed area.
In order to perform concave obstacle avoidance, we propose an hybrid DS capable of leveraging on either geodesic or harmonic motion depending on the need
\begin{equation}
    \ddot{\mathbf{x}} = - \boldsymbol{\Xi} \td{\v{x}} - S \rbr{\bar{\psi}(\mathbf{x}, \dot{\mathbf{x}})} \m{G}^{-1}\rbr{\m K \v x + \m{D} \dot{\mathbf{x}}},
    \label{eqn:hybrid_dynamics}
\end{equation}
where $S(\cdot)$ is a generalized sigmoid function, as in~\Cref{eqn:sigmoid_fun}, to smoothly transition between harmonic and geodesic motion.
\Cref{fig:local_field_second_concave_hybrid,fig:local_field_second_concave_hybrid_extreme}, respectively for semicircle and horseshoe obstacle shape, show the behavior of the DS in~\Cref{eqn:hybrid_dynamics}.
The DS in~\Cref{eqn:hybrid_dynamics} exhibits geodesic behavior near to obstacle, when approaching it, $\sigma \rbr{\bar{\psi}(\mathbf{x}, \dot{\mathbf{x}})} \approx 0$.
When leaving the obstacle, thanks to the velocity dependency introduced before, the DS exhibits harmonic behavior, given that $\sigma \rbr{\bar{\psi}(\mathbf{x}, \dot{\mathbf{x}})} \approx 1$, allowing to reach the attractor without being "captured" by the local deformation of the space.

\section{Synthetic Example}
\label{sec:synthetic_example}

\begin{figure}[ht]
    \centering
    \subfloat[]{ \includegraphics[width=.22\textwidth]{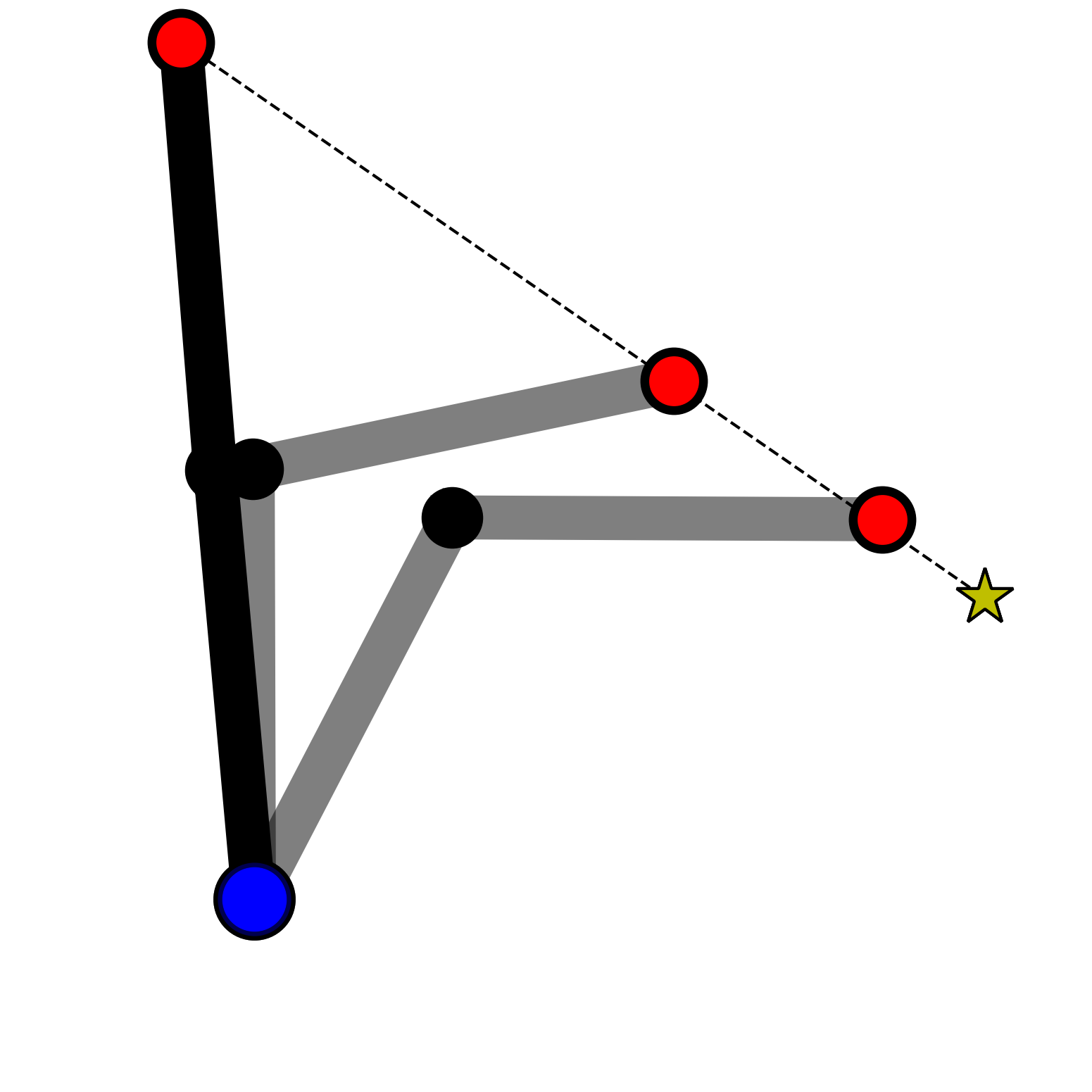}
        \label{fig:robot_motion}}
    \hfill
    \subfloat[]{\includegraphics[width=.24\textwidth]{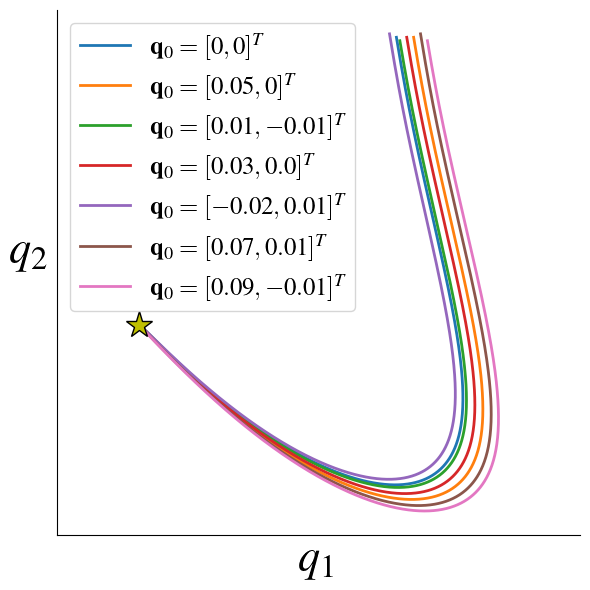}
        \label{fig:robot_trajectories}}
    \caption{\footnotesize (a) $2$-joints planar robotic arm performing a straight point-to-point motion of the end-effector; (b) corresponding trajectories in configuration space starting from different initial states.}
    \label{fig:synthetic_example}
\end{figure}

\begin{figure*}[ht]
    \includegraphics[width=\textwidth]{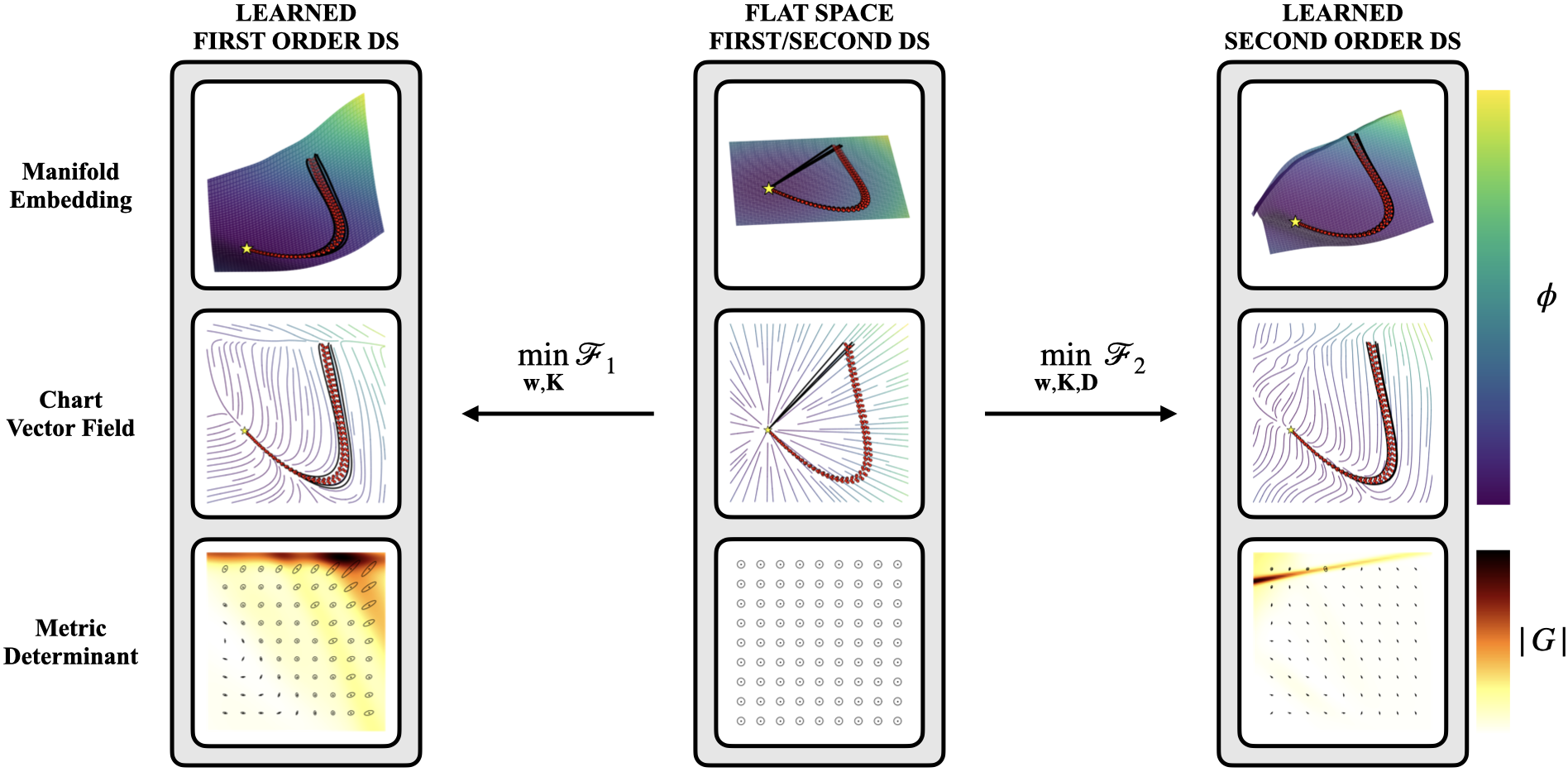}
    \caption{\footnotesize From top to bottom $3$D embedded representation of the latent manifold, chart space representation of the DS field induced by the manifold curvature and metric tensor's determinant and ellipsoids for selected locations: (center) before learning; (left) learned first-order DS; (right) learned second-order DS.}
    \label{fig:learning_ds}
\end{figure*}

In order to gain intuition on how the proposed method operates, in this Section, we start by analyzing a synthetic dataset achieved by gathering configuration space non-linear motions of a $2$-joints planar robotic arm.

Consider the 2-joints planar robotic arm in~\Cref{fig:robot_motion} whose state is represented by the vector $\mathbf{q} = [q_1, q_2]^T$.
Consider the generic equation of motion for a robotic arm, $\m{M}(\mathbf{q})\ddot{\mathbf{q}} + \m{C}(\mathbf{q}, \dot{\mathbf{q}})\dot{\mathbf{q}} = -\v{g}(\dot{\mathbf{q}}) + \boldsymbol{\tau}$, with $\m{M}(\mathbf{q})$, $\m{C}(\mathbf{q}, \dot{\mathbf{q}})$, $\v{g}(\dot{\mathbf{q}})$ and $\boldsymbol{\tau}$ being the inertia matrix, the Coriolis matrix, the gravity forces and input torques, respectively.
We notice that classical mechanical systems are Riemannian geometries with the inertia matrix, $\m{M}(\mathbf{q})$, playing the role of the metric tensor, $\m{G}(\mathbf{q})$, and the fictitious (or Coriolis) forces, $\m{C}(\mathbf{q}, \dot{\mathbf{q}})$, representing the geometric forces derived by the product of the Christoffel symbols and the velocities, $\m{\Xi} \rbr{\v x, \td{\v x}}$.
We elicit the non-linear dynamics of the robot by generating straight motions in the task (end-effector) space using operation space control, $\boldsymbol{\tau}=-\m{J}(\mathbf{q})^T \left( \m{K}(\mathbf{x}-\mathbf{x}^*) + \m{D} \dot{\mathbf{x}} \right)$, where $\m{J}(\mathbf{q})$ is the Jacobian matrix relating joint space velocities to task space velocities, $\mathbf{x}^*$ the equilibrium point in task space and $\m{K}$ and $\m{D}$ tunable gain matrices.
Starting from different initial configurations, $\mathbf{q}_0$, we generate in total $7$ trajectories, \Cref{fig:robot_trajectories}, of which $4$ are used for training and $3$ for testing.

In order to learn the non-linear trajectories shown in~\Cref{fig:robot_trajectories}, we approximate $\psi$, in~\Cref{eqn:embedding}, with a feed-forward network composed by $2$ hidden layers of $32$ neurons each with hyperbolic tangent activation function to guarantee at least $C^1$ regularity; see~\Cref{app:ablation_study} for details.
The network's weights are randomly initialized very close to zero.
This yields an almost flat manifold in the embedding space representation, \Cref{fig:learning_ds} central column in the top, with the metric tensor approaching the identity Euclidean metric.
At the bottom of the central column, this is shown by the determinant of $\m{G}$, the color gradient in background, and the ellipses representing the principal direction of the metric tensor.
The determinant of the metric gives an absolute value of the local deformation of the space.
In this case, the determinant of  $\m{G}$ is constant and almost equal to $1$ everywhere.
In order to have an intuition of the direction of the deformation, the ellipses generated by the eigen decomposition of the metric tensor are shown for selected location.
\begin{figure}[ht]
    \centering
    \subfloat[]{\includegraphics[width=.24\textwidth]{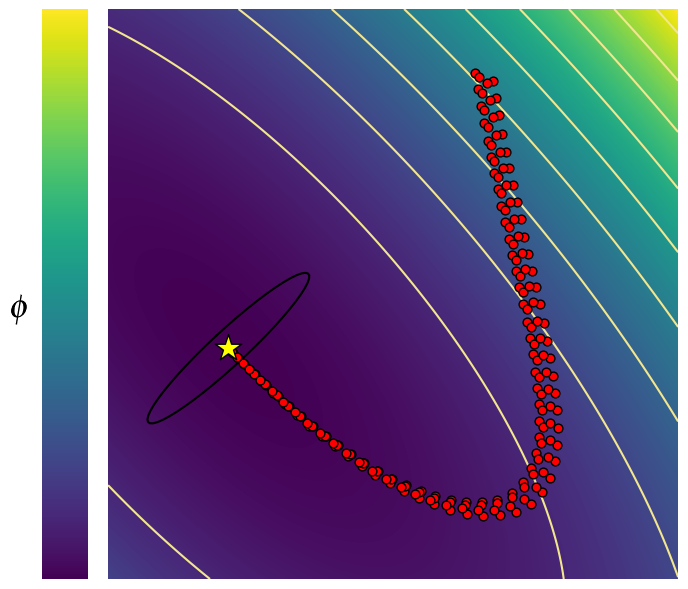}
        \label{fig:synthetic_first_potential}}
    \hfill
    \subfloat[]{\includegraphics[width=.205\textwidth]{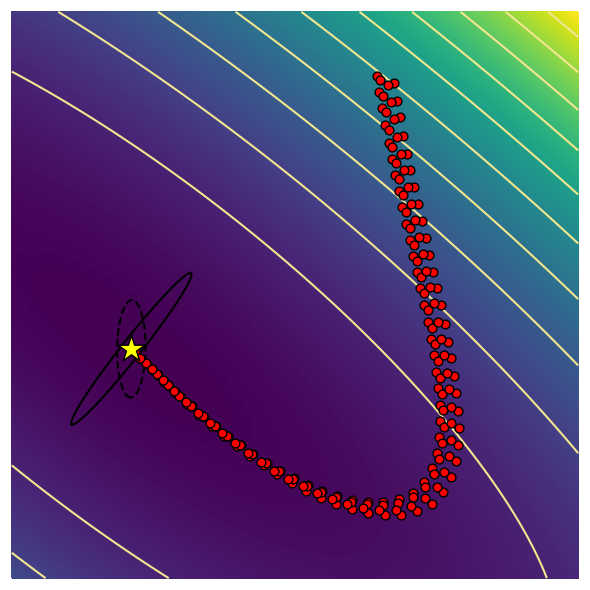}
        \label{fig:synthetic_second_potential}}
    \caption{\footnotesize Learned potential function with stiffness and dissipation matrices principal direction ellipsoids for (a) first-order and (b) second-order DS.}
    \label{fig:synthetic_second}
\end{figure}

\begin{figure*}[t]
    \centering
    \subfloat[]{\includegraphics[width=.19\textwidth]{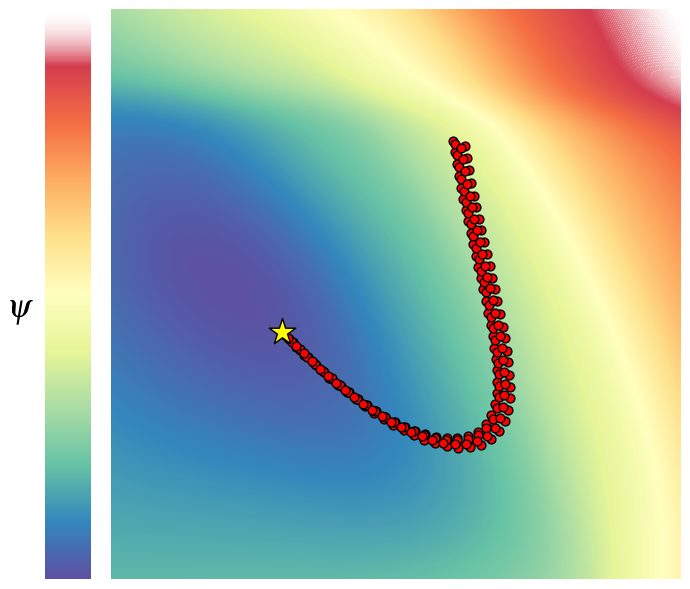}
        \label{fig:analysis_embedding_nobump}}
    \hfill
    \subfloat[]{\includegraphics[width=.19\textwidth]{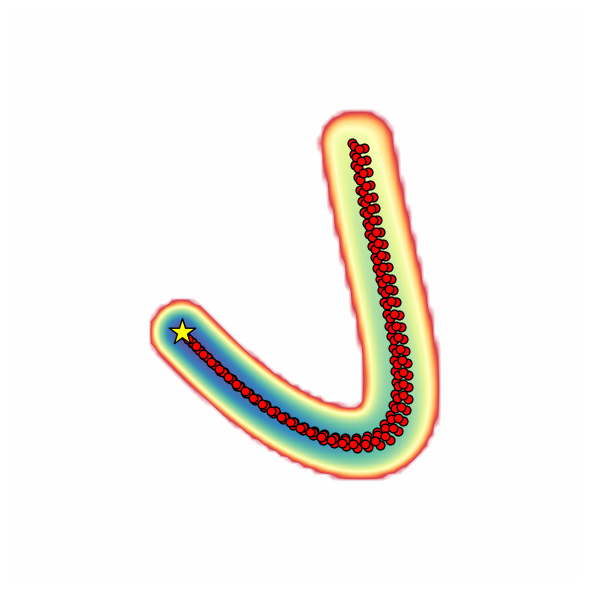}
        \label{fig:analysis_embedding_bump}}
    \hfill
    \subfloat[]{\includegraphics[width=.19\textwidth]{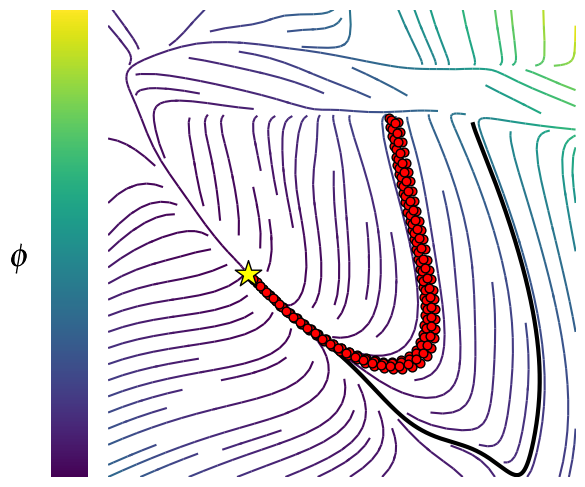}
        \label{fig:analysis_field_nobump}}
    \hfill
    \subfloat[]{\includegraphics[width=.155\textwidth]{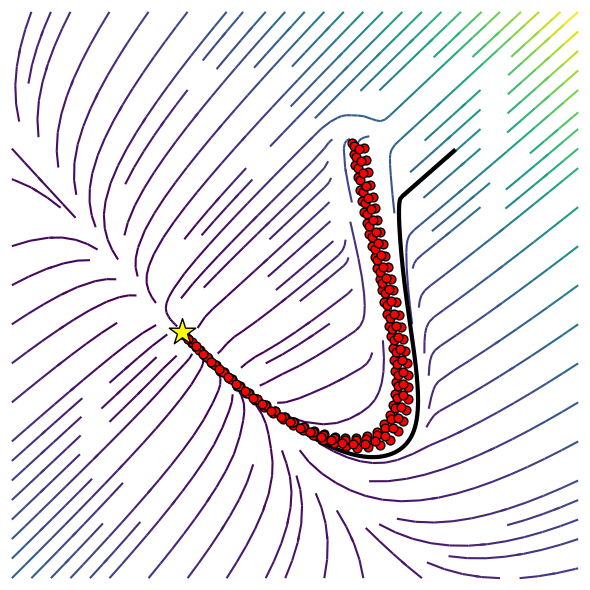}
        \label{fig:analysis_field_bump}}
    \hfill
    \subfloat[]{
        \includegraphics[width=.19\textwidth]{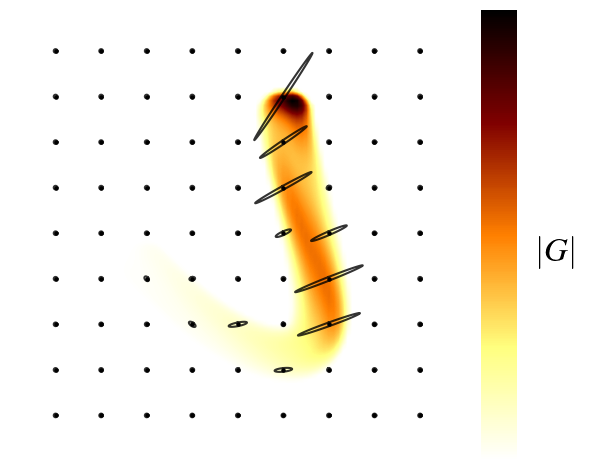}
        \label{fig:analysis_detmetric_bump}
    }
    \caption{\footnotesize (a) Gradient of the d$+1$ embedding component; (b) gradient of the d$+1$ embedding component with bump function; (c) one sampled trajectory with initial position far away from the demonstrated trajectories; (d) one sampled trajectory with initial position far away from the demonstrated trajectories with bump function; (e) metric determinant with eigenvalue decomposition ellipses of the metric from selected location.}
    \label{fig:analysis_embedding}
\end{figure*}

For the Euclidean metric generated by the almost flat space condition, such ellipses approach the shape of a circle.
For the first order DS learning, we start from a spherical stiffness matrix.
For the second order DS, we additionally set the damping matrix to yield an initial critically damped behavior in flat space, $\m{D} = 2(\m{K}\m{M})^{1/2}$.
These conditions, for both DSs, results in a linear vector field, in the chart space representation, where the sampled streamlines are straight lines towards the attractor, in the middle of the central column in~\Cref{fig:learning_ds}.
Note that, for the second-order DS, the vector field is obtained by integrating one step forward~\Cref{eqn:full_ds}, considering an initial velocity of zero.
During the training process, the stiffness and the damping matrices can be spherical, diagonal or Symmetric Positive Definite (SPD), as well as fixed and therefore not considered as an optimization variable.
The stiffness and dissipation matrices, together with the curvature of the manifold, contribute to the non-linearity of the learned DS.

\Cref{fig:synthetic_first_potential,fig:synthetic_second_potential} show the isoline of the potential function, after the learning, for the first and second order DS.
In this example, we opted for an SPD matrix for both the stiffness and the damping matrices.
Observe that in both cases, first and second order DS, the learnt stiffness matrix "aligns" itself orthogonally to the direction of demonstrated trajectories in the neighborhoods of the attractor.
Therefore, part of the contribution to the non-linearity of the streamlines is outsourced to the potential function gradient.
The remaining non-linearity needed for learning the demonstrated DS is taken over by the curvature of the space.

The top row of~\Cref{fig:learning_ds} shows the embedded representation of the learnt manifold for the first (left column) and second (right column) order DS.
In the bottomr line, the behavior of the determinant of the metric for both DSs.
In the case of the first-order DS, the curvature of the space gives rise to an \emph{energetic barrier} at the onset of the trajectory, guiding the flow downward.
In the lower-right region of the space, it steers the streamlines towards the attractor.
This phenomenon is further elucidated by examining the principal directions indicated by the ellipses of the metric tensor.
The ellipses experience a compression perpendicular to the direction of maximal deformation.
As a consequence, this leads to a projection of DS velocities tangentially to the deformation of the space, resulting in the desired non-linearity.
In the case of the second-order DS shown at the bottom, we encounter a similar scenario, but this time with a more pronounced energetic barrier in the lower region of the space.
This barrier, via the Christoffel symbols, induces directional deceleration, effectively guiding the streamlines towards the attractor.

The resulting chart space representation of the vector field along with $3$ sampled testing trajectories is shown in the central row.
The underlying deformation of the space induces an apparent non-linearity in the chart space representation of both DSs.
The streamlines curve, adapting to the shape of the demonstrated trajectory for the first-order DS.
Differently, in the second-order DS, the sampled streamlines do not evolve accordingly to the background vector field.
Indeed, for the second-order DS, the background vector field assumes in each point zero initial velocity.

\begin{figure}[ht]
    \centering
    \subfloat[]{
        \includegraphics[width=.22\textwidth]{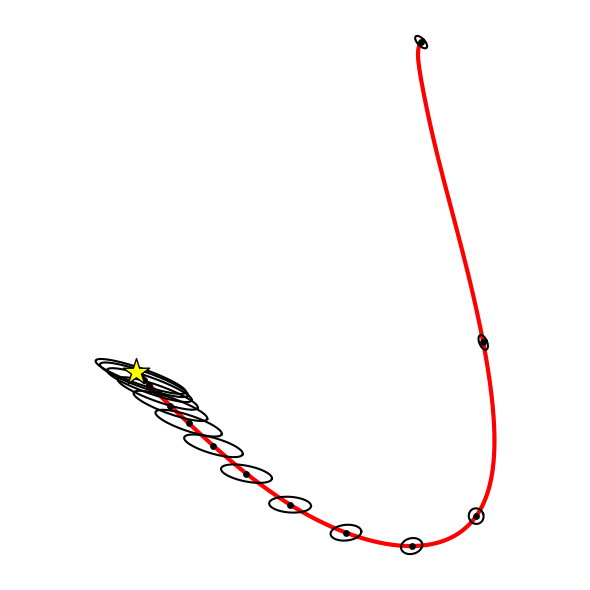}
        \label{fig:synthetic_metric_first}
    }
    \hfill
    \subfloat[]{
        \includegraphics[width=.22\textwidth]{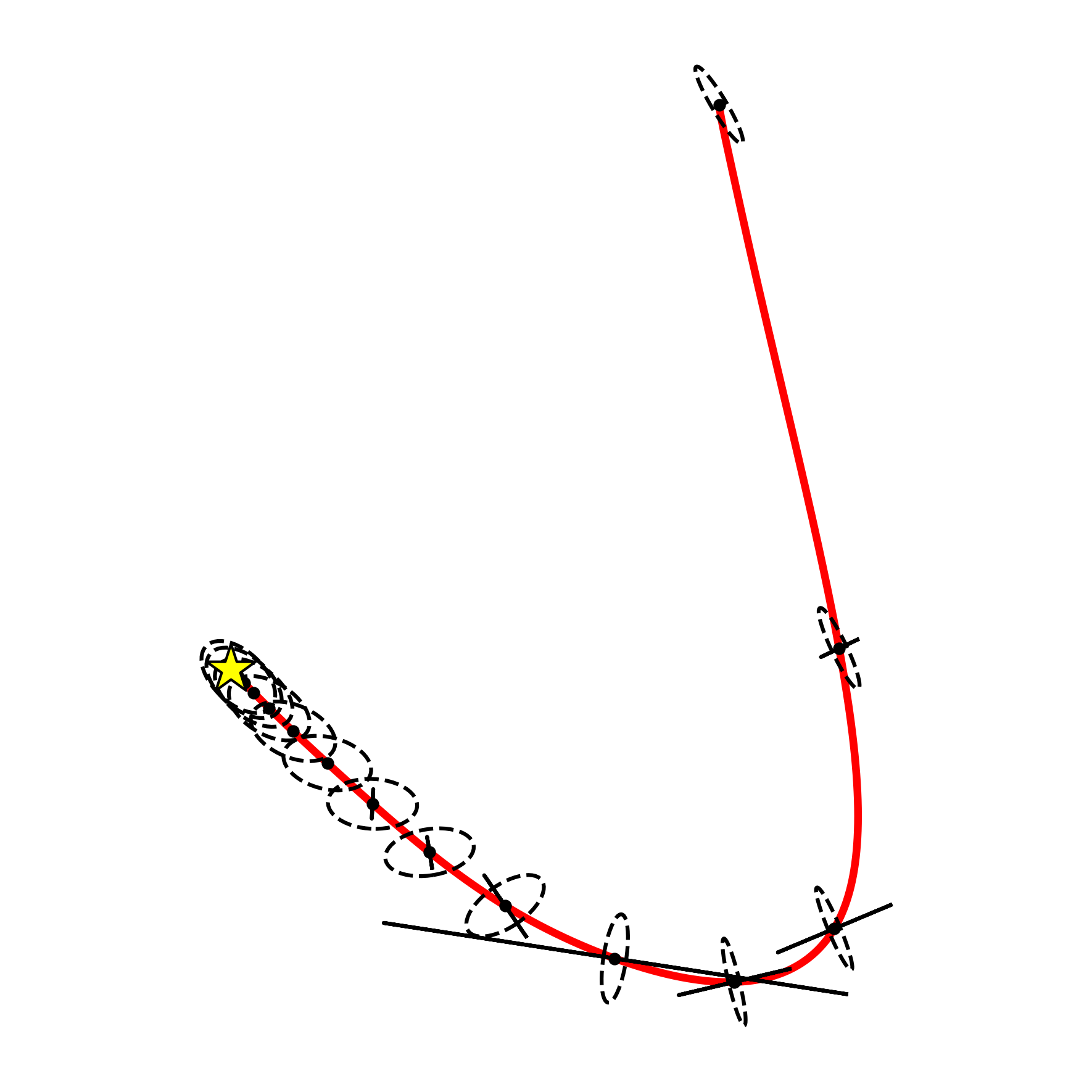}
        \label{fig:synthetic_christoffel_second}
    }
    \caption{\footnotesize Evolution of (a) inverse metric and (b) Christoffel symbols principal directions along one sampled streamline.}
\end{figure}
Consider the first order DS. \Cref{fig:synthetic_metric_first} shows how the principal directions of the inverse metric tensor vary along the one sampled trajectory due to deformation of the space.
In the first part of the trajectory the ellipses are considerably elongated in $q_1$.
By projecting the DS current velocity onto the inverse metric principal axis, see~\Cref{eqn:first_ds}, the velocity of the DS increases in the direction of $q_1$, generating the non-linear behavior depicted.
Similar consideration, \Cref{fig:synthetic_christoffel_second}, can be done for the second-order DS by analyzing the Christoffel symbols (solid line) and metric tensor inverse (dashed line) ellipses.

\begin{figure*}[t]
    \centering
    \subfloat[]{\includegraphics[width=.18\textwidth]{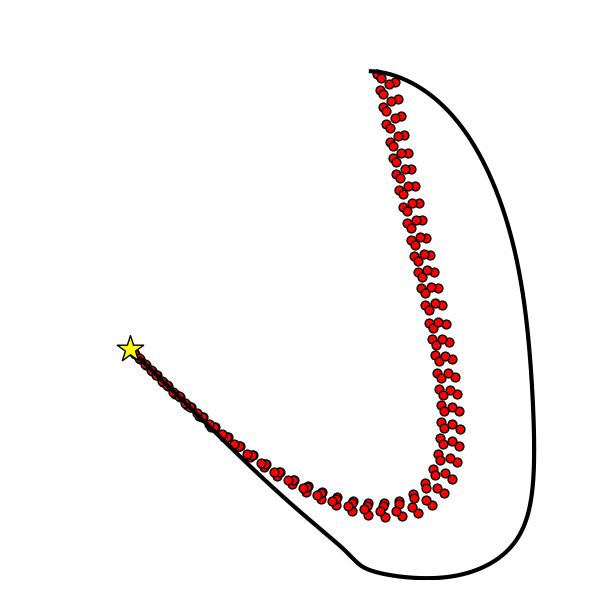}
        \label{fig:synthetic_field_second_nodirdissipation}}
    \hfill
    \subfloat[]{\includegraphics[width=.18\textwidth]{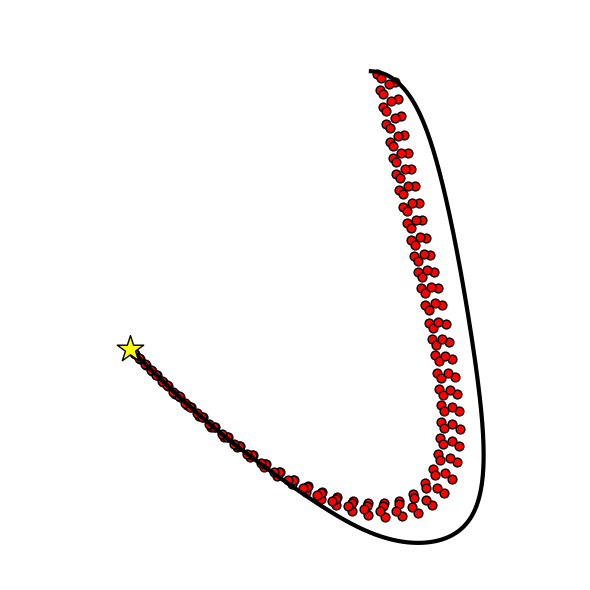}
        \label{fig:synthetic_field_second_dirdissipation10}}
    \hfill
    \subfloat[]{\includegraphics[width=.18\textwidth]{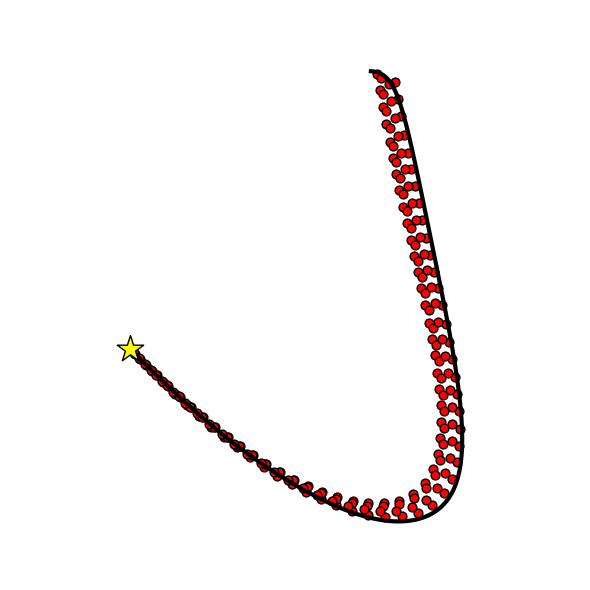}
        \label{fig:synthetic_field_second_dirdissipation20}}
    \quad
    \subfloat[]{\includegraphics[width=.18\textwidth]{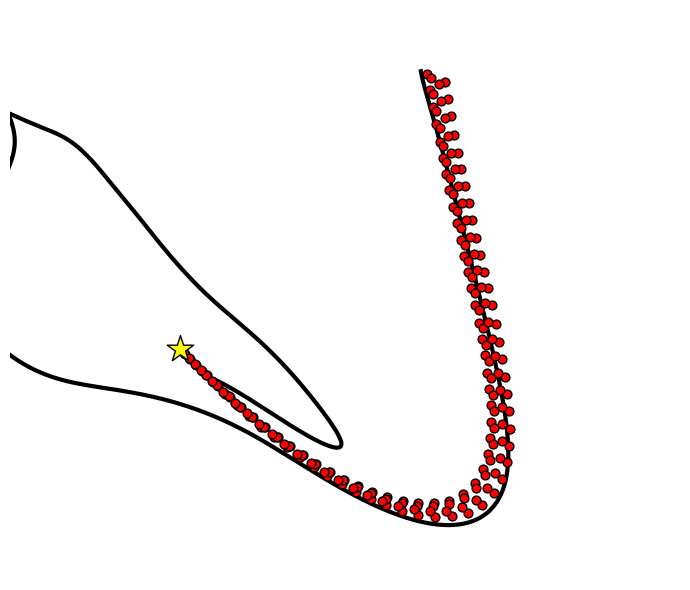}
        \label{fig:synthetic_field_second_overshoot}}
    \hfill
    \subfloat[]{\includegraphics[width=.18\textwidth]{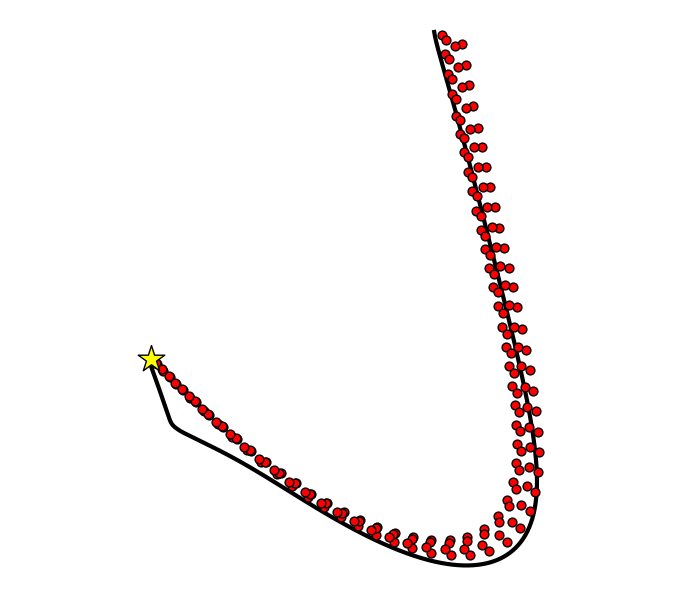}
        \label{fig:synthetic_field_second_expdissipation}}
    \caption{\footnotesize Second-order DS (a) without, (b)-(c) with ($\lambda=10$ and $\lambda=20$) directional dissipation. Second-order DS (d) without and (e) with exponential dissipation.}
    \label{fig:field_second_dirdissipation}
\end{figure*}

Notice that the DS is linear on the manifold and follows a mass-spring-damper system as in~\Cref{eqn:vector_ds}.
The curvature of the manifold makes the DS appear non-linear in the chart (Euclidean) space representation of the manifold.

\subsection{Locally Active Space Deformation}
\label{app:locally_active_deformation}
Despite global stability guarantees, when using global function approximator such as neural networks, the behavior far away from demonstrations is not predictable. \Cref{fig:analysis_embedding_nobump} shows the contour of the $d$+1 embedding component, approximated by a neural network, that directly influences the curvature of the manifold.
Close to the demonstrated trajectories, the function approximator ensures that the manifold's curvature would yield the desired DS non-linearity.
Away from the demonstrated trajectories, the local curvature of the manifold produces a DS behavior that can be sub-optimal or even undesired, see~\Cref{fig:analysis_field_nobump}.

In order to alleviate this problem, we propose to flatten the space far away from the demonstrations.
This is operation is performed at query time and it does not influence the training process.
In this case, by flattening the space away from the demonstrations, the DS will exhibit global linear behavior except for the regions where training points are available.
The nominal behavior far away from the demonstration is not limited to be linear. By choosing different types of nominal curvature, the DS will exhibit different behaviors.

To enforce flat space behavior in those areas where training data is not available, we deploy a distance dependent bump function.
Let $\s{X}$ denoting the training set.
We define
\begin{equation}
    \psi_{\t{bumped}} \rbr{\v{x}; \v{w}} = \alpha(\v{x}) \psi\rbr{\v{x}; \v{w}},
\end{equation}
where $\alpha(\v{x})$ is a bump function defined as
\begin{equation}
    \alpha(\v{x}) =
    \begin{cases}
        \frac{1}{e}\exp\rbr{-\frac{r^2}{r^2-\f{dist}(\v{x}, \s{X})^2}}, & \f{dist}\rbr{\v{x}, \s{X}} \le r \\
        0,                                                              & \t{otherwise}.
    \end{cases}
\end{equation}
$\f{dist}\rbr{\v{x}, \s{X}}$ is the distance between $\v{x}$ and its nearest neighborhood $\v x_i \in \s{X}$ \footnote{In oder to increase regularity of the $\psi_{\t{bumped}} \rbr{\v{x}; \v{w}}$, $\f{dist}\rbr{\v{x}, \s{X}}$ can be computed as the average distance between $\v x$ and its $K$ nearest neighborhoods. This is still computationally cheap by adopting modern efficient implementation of KNN. Distribution based bump function may represent an alternative way to KNN. In this case the distribution can be learned offline (for instance via GMM) yielding almost zero cost at query time. In addition joint position-velocity distribution could be learned to enforce flat-space behavior away from the demonstrated velocities.}.
$r$ is a user-defined parameter that regulates how far from the demonstrated trajectories the manifold should start to have nominal zero curvature.

\Cref{fig:analysis_embedding_bump} shows the third embedding component behavior when pre-multiplied by the bumped function.
The manifold region in the neighborhood of the demonstration preserves its original curvature given by the learning procedure.
Away from the demonstration the manifold becomes increasingly flat.
This type of manifold structure yields a linear behavior away from the demonstrations that smoothly transitions towards nonlinear behavior when approaching the area of the demonstrated trajectories, see~\Cref{fig:analysis_field_bump}.
By analyzing the determinant of the metric tensor, \Cref{fig:analysis_detmetric_bump}, we can clearly notice how this type of embedding structure affects the curvature only in localized portion of the space.
The ellipses of the metric tensor away from the demonstrated trajectories converge to circles; constant and equal eigenvalues of value $1$.
Close to the demonstrations the ellipses deform, yielding the correct geometric accelerations to follow the demonstrated trajectories.

\subsection{Directional \& Exponential Dissipation}
Second-order DSs offer a richer and more versatile way to articulate high-level policies compared to first-order DS.
However, their application in LfD is uncommon.
Primarily, user demonstrations tend to emphasize position over velocity.
For instance, during a kinaesthetic demonstration with a robot---manually guiding the robot's end-effector along a desired path---focus is primarily on the sequence of positions rather than the end-effector's velocity.
Consequently, this oversight can result in unintended behaviors near the demonstrated positions, especially when the state velocity differs from the demonstrated one.
In this section, we present two potential strategies aimed at mitigating this issue when deploying second-order DSs.

\begin{figure*}[t]
    \centering
    \subfloat[]{\includegraphics[width=.18\textwidth]{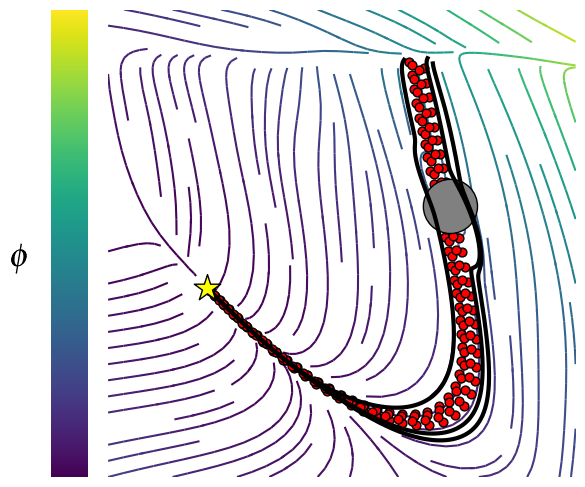}
        \label{fig:synthetic_first_field_obstacle_classic}}
    \subfloat[]{\includegraphics[width=.15\textwidth]{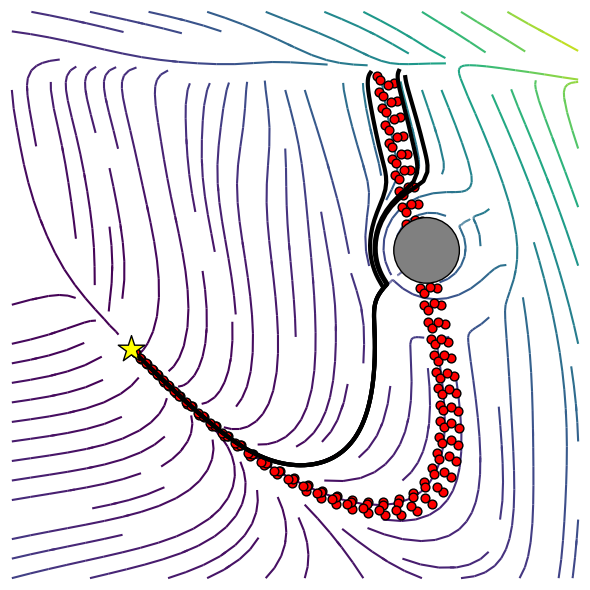}
        \label{fig:synthetic_first_field_obstacle_direct}}
    \subfloat[]{\includegraphics[width=.15\textwidth]{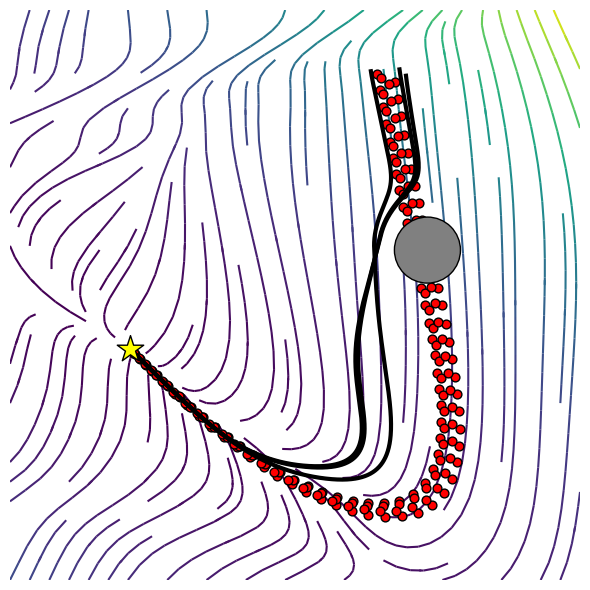}
        \label{fig:synthetic_second_field_obstacle_direct}}
    \subfloat[]{\includegraphics[width=.15\textwidth]{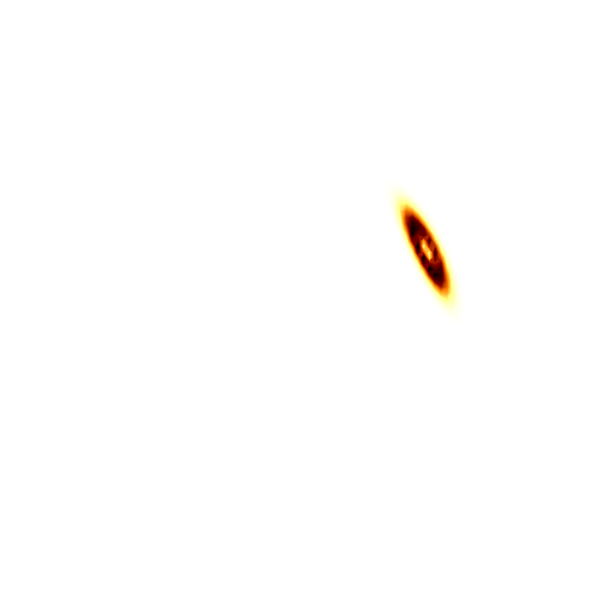}
        \label{fig:synthetic_first_detmetric_obstacle_classic}}
    \subfloat[]{\includegraphics[width=.15\textwidth]{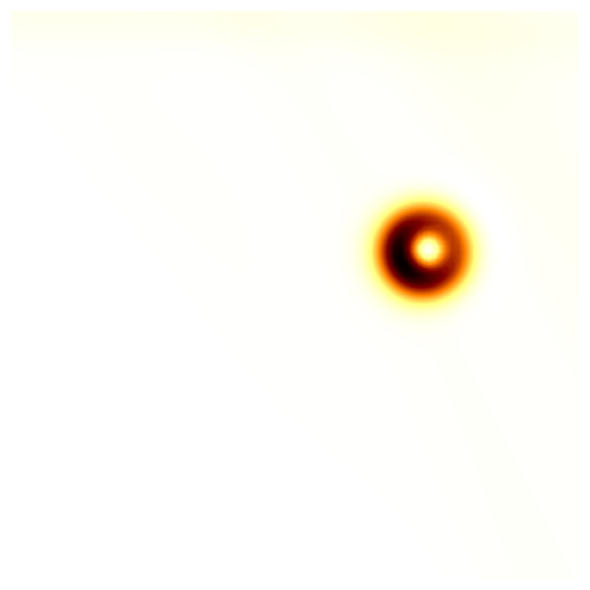}
        \label{fig:synthetic_first_detmetric_obstacle_direct}}
    \subfloat[]{\includegraphics[width=.18\textwidth]{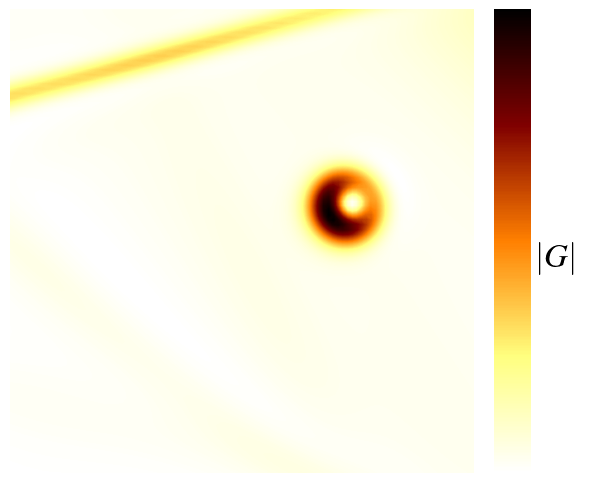}
        \label{fig:synthetic_second_detmetric_obstacle_direct}}
    \caption{\footnotesize Vector field and metric determinant for (a)-(d) first-order DS with classic method; (b)-(e) first-order DS with direct deformation method; (c)-(f) second-order DS with direct deformation method.}
    \label{fig:learning_ds_obs_first}
\end{figure*}

\Cref{fig:synthetic_field_second_nodirdissipation} shows the streamline sampled from the initial position of one of the testing trajectories.
In this case the initial velocity is not equal to zero, differently from what we had during the training.
As it is possible to see in this scenario, the sampled streamline does not follow at all the demonstrations taking an alternative path to reach the attractor.
When controlling, for instance, on real robot's end-effector, we cannot ensure the current velocity of the end-effector will always lie to the demonstrated ones for each specific position, especially when the controlled system has to answer to compliance requisite in the interaction with humans.
In order to make second-order DS reliable in this type of scenario and alleviate undesired behaviors whenever we have initial velocities far away from the demonstrated ones, we propose to add, without loss of stability, an additional directional dissipation
\begin{align}
    \ddot{\mathbf{x}} & = -\m{G}^{-1}(\mathbf{x};\mathbf{w}) \left(\m{K}(\mathbf{x} - \mathbf{x}*) + \m{D} \dot{\mathbf{x}} \right) - \boldsymbol{\Xi} (\mathbf{x},\dot{\mathbf{x}};\mathbf{w}) \dot{\mathbf{x}} \notag \\
                      & - \lambda_{\text{dir}}(\frac{\td{\v x}}{\norm{\td{\v x}}} - \frac{\td{\v x}*}{\norm{\td{\v x}*}}).
\end{align}
$\td{\v x}*$ is a first-order reference DS, providing the desired velocity field nominal behavior.
Whenever first-order DS following the demonstrated trajectories is available we can take advantage of it in order to steer our second-order DS within the demonstrated velocities.
\Cref{fig:synthetic_field_second_dirdissipation10,fig:synthetic_field_second_dirdissipation20} show the resulting streamline for increasing values of $\lambda_{\text{dir}}$.

Another issue encountered in adopting second-order DS is the undesired under-damped behavior shown in \Cref{fig:synthetic_field_second_overshoot}.
If for classical harmonic damped oscillator we can easily set the damping term so to have a critically-damped behavior, this is not possible for oscillators on manifolds whereas the Christoffel symbols acts as a non-linear damping term not under our control.
This leads to trajectory overshoot in the attractor area.
In order to alleviate this problem, an additional dissipative force can be considered
\begin{align}
    \ddot{\mathbf{x}} & = -\m{G}^{-1}(\mathbf{x};\mathbf{w}) \left(\m{K}(\mathbf{x} - \mathbf{x}*) + \m{D} \dot{\mathbf{x}} \right) - \boldsymbol{\Xi} (\mathbf{x},\dot{\mathbf{x}};\mathbf{w}) \dot{\mathbf{x}} \notag \\
                      & - \lambda_{\text{exp}} \exp \rbr{- \tau \norm{\v x - \v x*}^2}.
\end{align}
This new dissipative force acts only locally and it grows exponentially approaching the attractor.
As shown in \Cref{fig:synthetic_field_second_expdissipation} this term effectively arrest the DS streamline at the attractor removing the problem of overshooting.

\subsection{Obstacle Avoidance Online Direct Deformation}
We now show how the learned DS can adapt to the presence of obstacles.
As analyzed in~\Cref{sec:obstacle_avoidance} we have two ways of performing obstacle avoidance leveraging on the geometrical structure of the space.

The first method, as in \cite{beik-mohammadi_learning_2021}, consists in designing a specific metric that acts in the ambient space.
All the geometric terms are then derived as shown in~\Cref{sec:learning_embedding} with $\m H = \hat{\m{H}} \neq \m I$.
We will refer to this approach as \emph{classical method}.

In our framework, we have the additional and more intuitive option of directly deforming the space by altering locally the embedding map post-training, $\hat{\psi}(\mathbf{x};\bar{\mathbf{w}}) = \psi(\mathbf{x};\bar{\mathbf{w}}) + \bar{\psi}(\mathbf{x})$.
We will refer to our approach as \emph{direct deformation method}.

For the classical method we opt, as barrier function, for the Gaussian kernels given by $k(\bar{\mathbf{y}}, \mathbf{y}) = \text{exp} \left( -\frac{\norm{\bar{\mathbf{y}} - \mathbf{y}}^2}{2\sigma^2} \right)$, with kernel width $\sigma$. $\bar{\mathbf{y}} = \mathbf{\Psi}(\bar{\mathbf{x}})$ is the obstacle position in the ambient space.
The ambient metric is then constructed as in~\Cref{eqn:ambient_metric}.

For the direct deformation of the space the same Gaussian kernel acting in the chart space $k(\bar{\mathbf{x}}, \mathbf{x})$ can be used, with no need to query the ambient space location of the obstacle.
The embedding is then altered as in~\Cref{eqn:embedding_obstacle}.

\begin{figure}[ht]
    \centering
    \subfloat[]{\includegraphics[width=.23\textwidth]{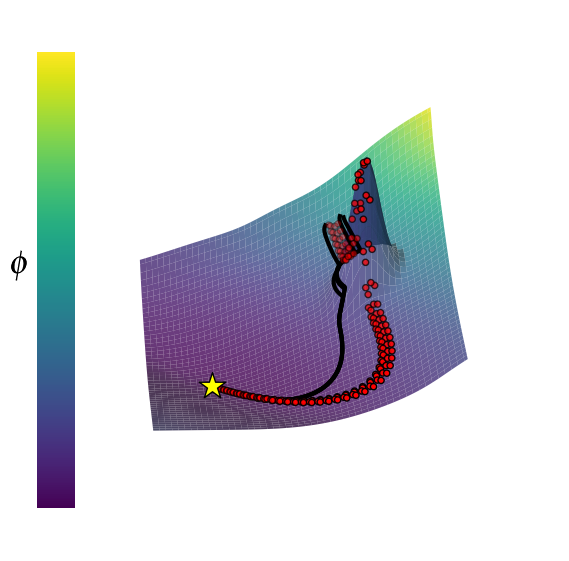}
        \label{fig:synthetic_first_obstacle_direct_embedding}}
    \subfloat[]{\includegraphics[width=.23\textwidth]{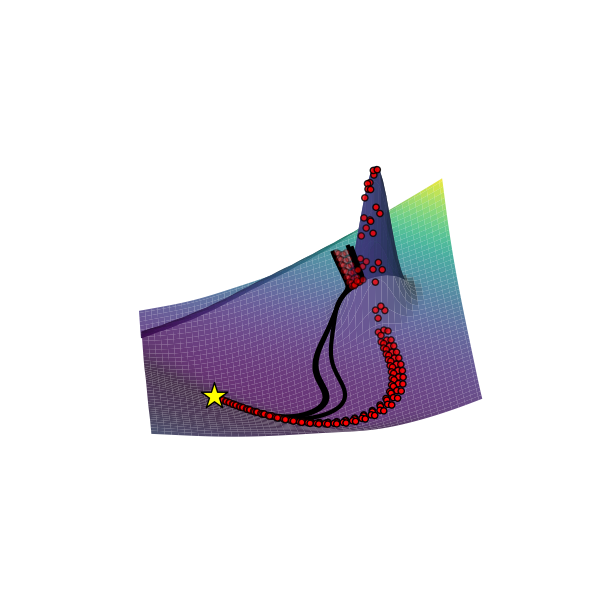}
        \label{fig:synthetic_second_obstacle_direct_embedding}}
    \caption{Embedding space visualization of the local deformation due to the obstacle presence for (a) first-order DS and (b) second-order DS.}
\end{figure}
$\sigma$ is a user-defined parameter regulating the speed of the decay of the local deformation.
One straightforward way of setting this parameter is imposing the desired decay at the border of the obstacle.
For instance $\sigma = \sqrt{-\frac{1}{2} \frac{r^2}{\t{log}\epsilon}}$, where $r$ is the local radius of the obstacle and $\epsilon$ is value of the kernel at $r$; typically $\epsilon \leq 1$e-3.
Barrier functions of the type $k(\bar{\mathbf{y}}, \mathbf{y}) = \text{exp} \left( \frac{a}{b(\norm{\bar{\mathbf{y}} - \mathbf{y}} - r)^b} \right)$ can be used as well. $a$ and $b$ are user-defined parameters that regulate the entity of the local deformation of the space and $r$ is the radius of the obstacle as before.

\begin{figure*}[t]
    \centering
    \subfloat{\includegraphics[width=.15\textwidth]{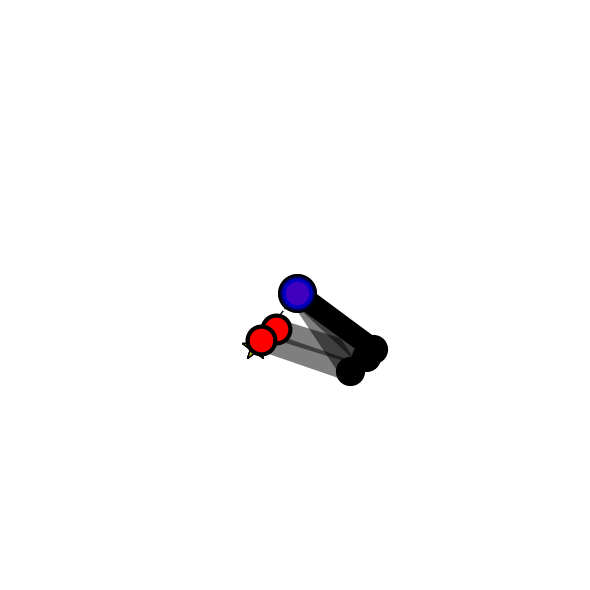}}
    \hfill
    \subfloat{\includegraphics[width=.15\textwidth]{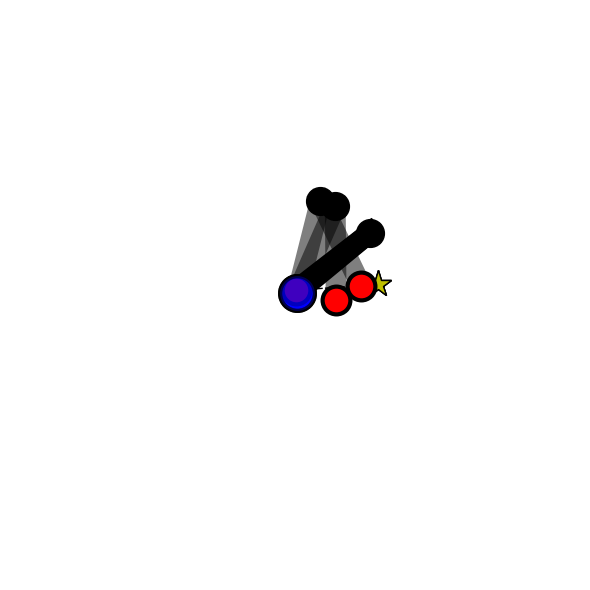}}
    \hfill
    \subfloat{\includegraphics[width=.15\textwidth]{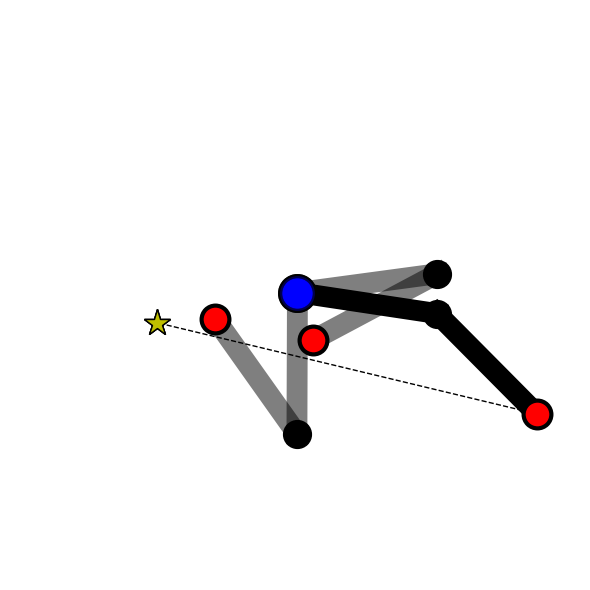}}
    \hfill
    \subfloat{\includegraphics[width=.15\textwidth]{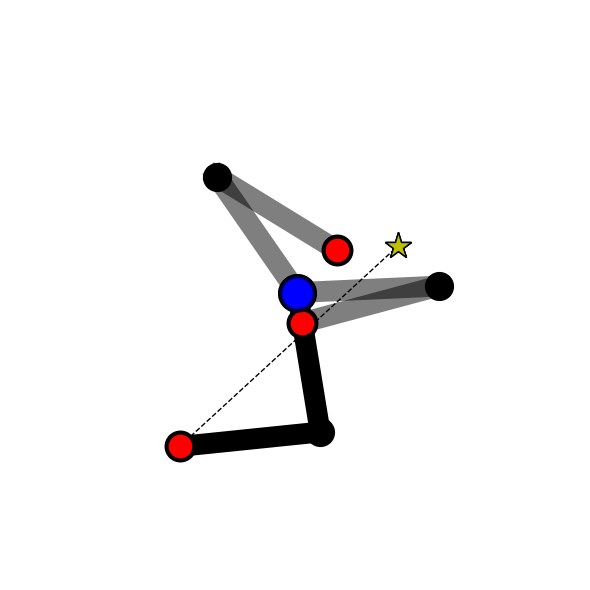}}
    \hfill
    \subfloat{\includegraphics[width=.15\textwidth]{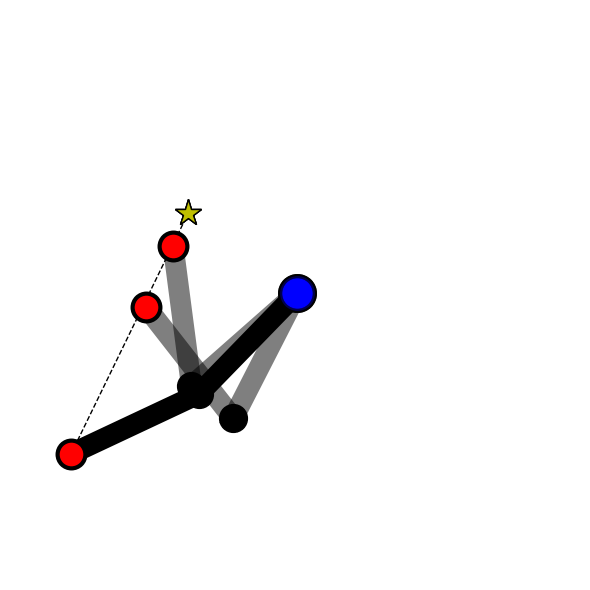}}
    \hfill
    \subfloat{\includegraphics[width=.15\textwidth]{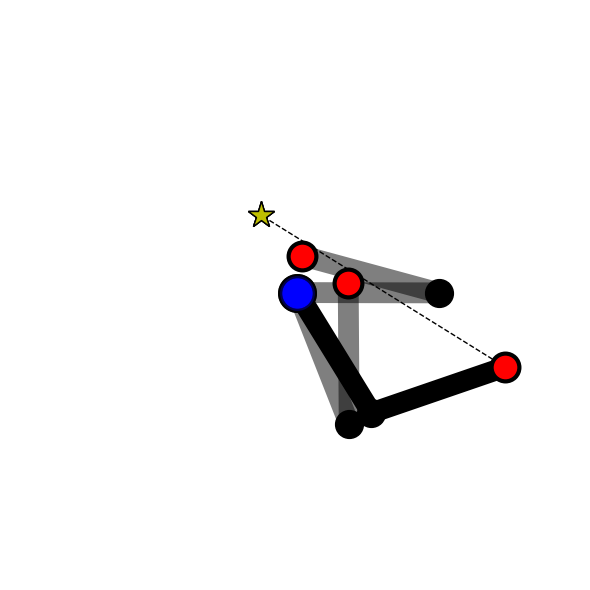}}
    \\
    \renewcommand{\thesubfigure}{a}
    \subfloat[]{\includegraphics[width=.15\textwidth]{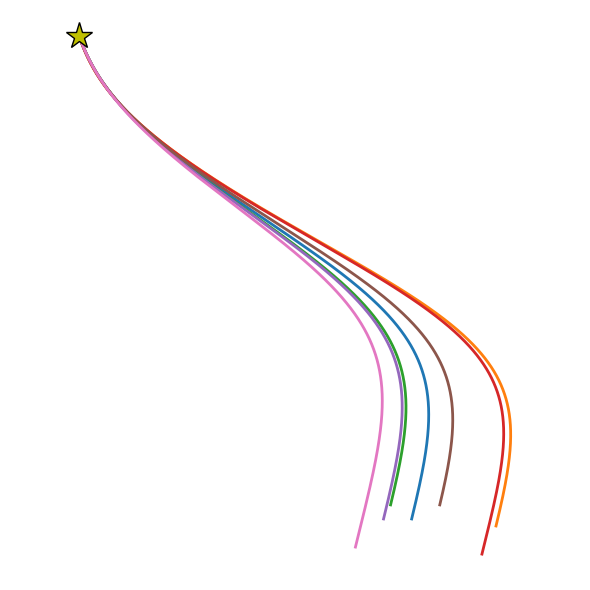}
        \label{fig:demo_1}}
    \hfill
    \renewcommand{\thesubfigure}{b}
    \subfloat[]{\includegraphics[width=.15\textwidth]{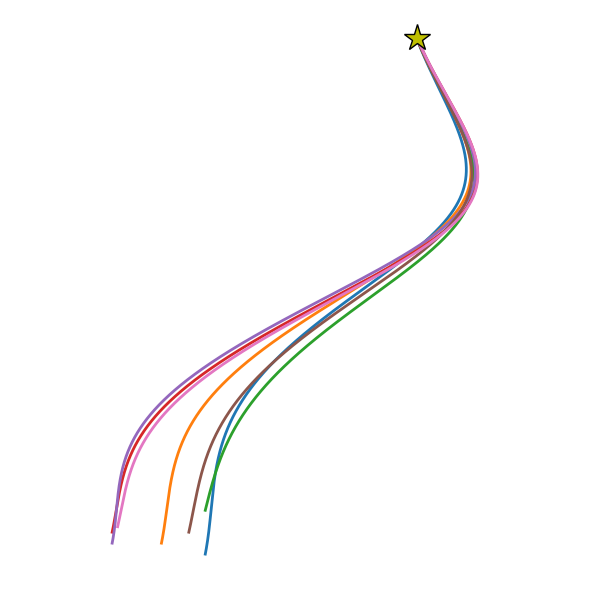}
        \label{fig:demo_2}}
    \hfill
    \renewcommand{\thesubfigure}{c}
    \subfloat[]{\includegraphics[width=.15\textwidth]{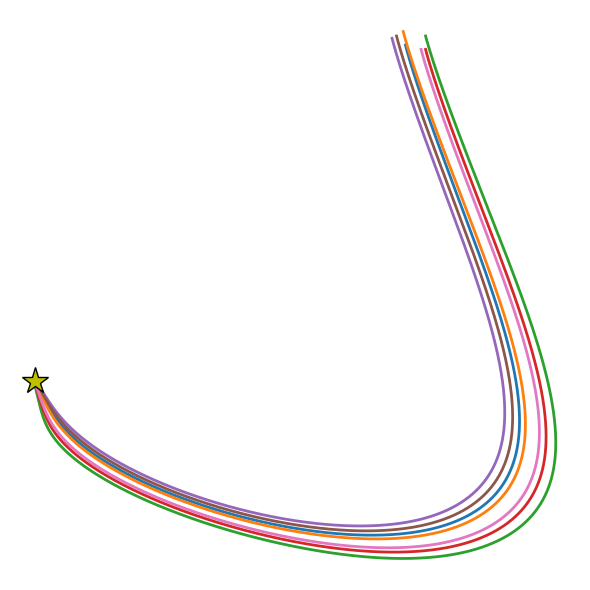}
        \label{fig:demo_3}}
    \hfill
    \renewcommand{\thesubfigure}{d}
    \subfloat[]{\includegraphics[width=.15\textwidth]{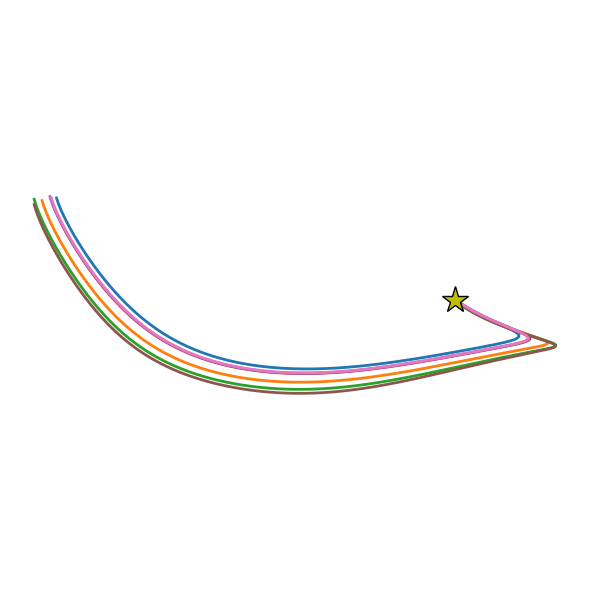}
        \label{fig:demo_4}}
    \hfill
    \renewcommand{\thesubfigure}{e}
    \subfloat[]{\includegraphics[width=.15\textwidth]{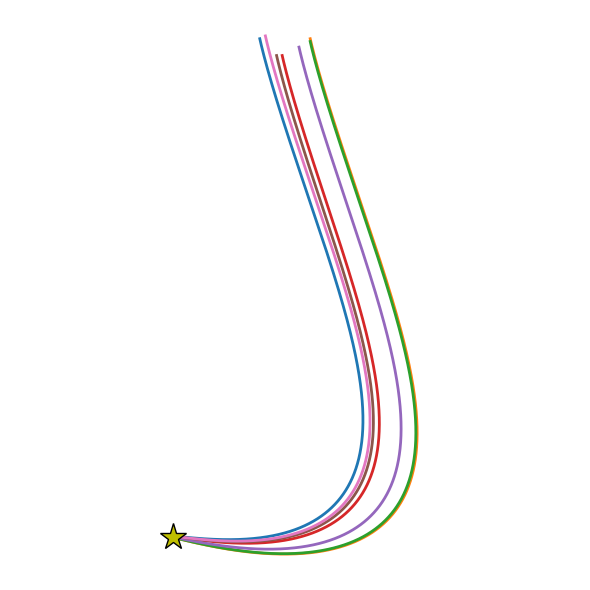}
        \label{fig:demo_5}}
    \hfill
    \renewcommand{\thesubfigure}{f}
    \subfloat[]{\includegraphics[width=.15\textwidth]{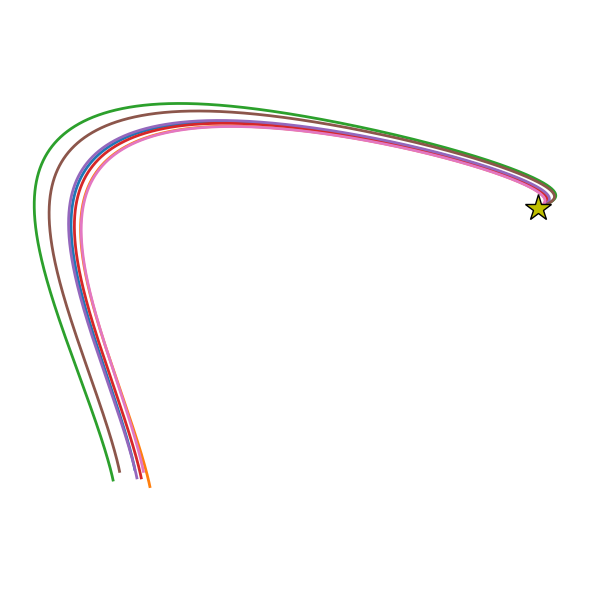}
        \label{fig:demo_6}}
    \caption{\footnotesize Evaluation: (top) Motion frames of 2-joints planar robotic manipulator; (bottom) sampled DSs in configuration space, corresponding to the above depicted robotic motion.}
    \label{fig:synthetic_results}
\end{figure*}

\begin{table*}[hb]
    \centering
    \resizebox{\textwidth}{!}{
\begin{tabular}{ccccccccccccc}
    \toprule
                           &                              & \multicolumn{3}{c}{\textbf{RMSE} [$\nicefrac{\text{rad}}{s}$]} &                     & \multicolumn{3}{c}{\textbf{CS} [$\text{rad}$]} &  & \multicolumn{3}{c}{\textbf{DTWD} [$\text{rad}^2$]}                                                                                                                \\
    \cline{3-5} \cline{7-9} \cline{11-13}                                                                                                                                                                                                                                                                                                                                \\
    \textbf{Demonstration} &                              & \textbf{Baseline}                                              & \textbf{$1$st - DS} & \textbf{$2$nd - DS}                            &  & \textbf{Baseline}                                  & \textbf{$1$st - DS} & \textbf{$2$nd - DS} &  & \textbf{Baseline} & \textbf{$1$st - DS} & \textbf{$2$nd - DS} \\
    \subref{fig:demo_1}    &                              &
    $1.35  \pm  0.66$      & $0.40  \pm  0.20$            & $\mathbf{0.02  \pm  0.00}$                                     &                     &
    $0.04  \pm  0.03$      & $0.01  \pm  0.00$            & $\mathbf{0.01  \pm  0.00}$                                     &                     &
    $1.04  \pm  0.89$      & $0.20  \pm  0.16$            & $\mathbf{0.15  \pm  0.06}$                                                                                                                                                                                                                                                                                   \\
    \subref{fig:demo_2}    &                              &
    $6.81  \pm  3.10$      & $0.30  \pm  0.07$            & $\mathbf{0.02  \pm  0.00}$                                     &                     &
    $0.37  \pm  0.25$      & $0.01  \pm  0.00$            & $\mathbf{0.01  \pm  0.00}$                                     &                     &
    $2.57  \pm  1.10$      & $\mathbf{0.28  \pm  0.07}$   & $0.70  \pm  0.24$                                                                                                                                                                                                                                                                                            \\
    \subref{fig:demo_3}    &                              &
    $ >10.00 $             & $ 1.31  \pm  0.13 $          & $ \mathbf{0.04  \pm  0.01} $                                   &                     &
    $ 0.49  \pm  0.44 $    & $ 0.01  \pm  0.00 $          & $ \mathbf{0.00  \pm  0.00} $                                   &                     &
    $ >10.00 $             & $ 1.59  \pm  0.31 $          & $ \mathbf{1.19  \pm  0.26} $                                                                                                                                                                                                                                                                                 \\
    \subref{fig:demo_4}    &                              &
    $ >10.00 $             & $ 1.25  \pm  0.07 $          & $ \mathbf{0.05  \pm  0.01} $                                   &                     &
    $ 0.71  \pm  0.31 $    & $ 0.02  \pm  0.00 $          & $ \mathbf{0.00  \pm  0.00} $                                   &                     &
    $ >10.00 $             & $ 3.15  \pm  0.57 $          & $ \mathbf{2.33  \pm  0.18} $                                                                                                                                                                                                                                                                                 \\
    \subref{fig:demo_5}    &                              &
    $ 4.17  \pm  2.79 $    & $ 0.72  \pm  0.11 $          & $ \mathbf{0.02  \pm  0.00} $                                   &                     &
    $ 0.03  \pm  0.01 $    & $ 0.00  \pm  0.00 $          & $ \mathbf{0.00  \pm  0.00} $                                   &                     &
    $ 1.21  \pm  0.48 $    & $ \mathbf{0.26  \pm  0.11} $ & $ 0.31  \pm  0.10 $                                                                                                                                                                                                                                                                                          \\
    \subref{fig:demo_6}    &                              &
    $ >10.00 $             & $ 0.88  \pm  0.11 $          & $ \mathbf{0.03  \pm  0.00} $                                   &                     &
    $ 0.49  \pm  0.33 $    & $ 0.01  \pm  0.0 $           & $ \mathbf{0.00  \pm  0.00} $                                   &                     &
    $ 7.45  \pm  5.28 $    & $ 5.17  \pm  5.65 $          & $ \mathbf{1.04  \pm  0.45} $                                                                                                                                                                                                                                                                                 \\
    \bottomrule
\end{tabular}
}
    \caption{Evaluation results: for each sampled DS in~\Cref{fig:synthetic_results} we evaluate 1st and 2nd order learning DS against the Baseline with respect to the three metrics RMSE, Cosine Similarity Kernel in velocity space, DTWD.}
    \label{tab:synthetic_results}
\end{table*}

The training process is carried out using $\m{H}=\m{I}$ and  $\hat{\psi}(\mathbf{x};\bar{\mathbf{w}}) = \psi(\mathbf{x};\bar{\mathbf{w}})$.
The ambient metric, for the classical method, and the embedding map, for the direct deformation method, will be locally modified at test time to take into account the presence of obstacles.

\Cref{fig:synthetic_first_field_obstacle_classic} shows, for the first order DS, the resulting vector field and test sampled trajectories, adopting the classical method for performing obstacle avoidance.
In this scenario, it is possible to notice how the streamline do not avoid the obstacle properly.
Moreover, by looking at~\Cref{fig:synthetic_first_detmetric_obstacle_classic}, we can notice how the space has been deformed in a asymmetric way.
Despite the isotropic kernel used, the local deformation of the space spans a tilted elliptic area.
This is due to the presence of prior curvature in the embedded space given by the learnt DS.
Whenever considerable curvature is present in the manifold, the lack of the coupling term, see~\Cref{eqn:direct_deformation_metric}, can lead to undesired behavior in the neighborhood of the obstacle.

By considering the obstacle as a direct local deformation of the manifold, we recover the expected behavior of the vector field in the neighborhood of the obstacle, \Cref{fig:synthetic_first_field_obstacle_direct}.
The metric determinant, \Cref{fig:synthetic_first_detmetric_obstacle_direct}, displays space deformation consonant with the isotropic kernel used.
The same happens for a second-order DS, \Cref{fig:synthetic_second_field_obstacle_direct,fig:synthetic_second_detmetric_obstacle_direct}.
Differently from the first-order DS, as already observed already in \Cref{sec:second_deformation}, we notice how the streamlines detach more quickly from the obstacle showing asymmetric behavior before and after the local deformation.

In this case, we can directly visualize how the presence of the obstacle affects the manifold's curvature, \Cref{fig:synthetic_first_obstacle_direct_embedding,fig:synthetic_second_obstacle_direct_embedding}.
The manifold structure is not re-learnt and it remains globally consistent with the embedded representation shown in~\Cref{fig:learning_ds}.
The obstacle affects the geometry only locally allowing the streamlines to follow the demonstrated trajectories away from the obstacle.

\subsection{Evaluation}
\label{sec:synthetic_eval}

In order to evaluate the performance of the learned DS we employ three metrics: 1) Root Mean Square Error, $\t{RMSE}=\sqrt{\frac{1}{M} \sum_{i=1}^{M} \norm{\td{\v x}_i^{\t{ref}} - \td{\v x}_i^\t{DS}}^2}$, 2) Cosine Similarity, $\t{CS} = \frac{1}{M} \sum_{i=1}^{M} \abs{1 - \frac{\innerp{\td{\v x}_i^{\t{ref}}}{\td{\v x}_i^\t{DS}}}{\norm{\td{\v x}_i^{\t{ref}}}\norm{\td{\v x}_i^\t{DS}}}}$, and 3) Dynamic Time Warping Distance, DTWD.
(1) and (2) are point-wise metrics that measure the similarity between two vector fields; in the first case, magnitude and direction are considered while, in the second case, only direction influences the score.

For the first order system we have $\td{\v x}_i^\t{DS} = \v f \rbr{\v x_i^{\t{ref}}}$.
In order to compare first and second order systems with metric (1) and (2), for the second order system, we sample one step forward from the learned DS with initial condition given by $\rbr{\v x_i^{\t{ref}}, \td{\v x}_i^\t{ref}}$. Using the same sampling frequency, $h$, of the testing trajectories we have
\begin{equation}
    \td{\v x}_{i+1}^\t{DS} =  \td{\v x}_i^\t{ref} + \frac{1}{h} \v f \rbr{\v x_i^{\t{ref}}, \td{\v x}_i^\t{ref}} \quad \t{for} \quad i = 1, \dots, N-1,
\end{equation}
where $N$ the number of sampled points per testing trajectory.

(3) measures the dissimilarity between the shape of a reference trajectory and its corresponding reproduction from the same initial points (and velocity for second order systems). In this case, for each testing trajectory, we sample a streamline starting from initial condition $\v x^{\t{ref}}_1$, for the first order DS, and $\rbr{\v x_1^{\t{ref}}, \td{\v x}_1^\t{ref}}$, for the second order DS, with sampling frequency $h$. The sampling frequency is coincident with the one used to sample the testing trajectories.

We compare our approach against the learning dynamical systems via diffeomorphism in \cite{rana_euclideanizing_2020}.
We refer to this approach as \emph{Euclideanizing Flows} (EF)\footnote{Efficient \textsf{PyTorch}-based implementation of EF available at:\\\url{https://github.com/nash169/learn-diffeomorphism}}.
This approach adopts an NVP transformation structure, \cite{dinh_density_2016}, where the diffeomorphism is achieved by a sequence of the so-called \emph{coupling layers}.
Each coupling layers is composed by operations of scaling and translation approximated by weighted sum of Random Fourier Features (RFF), \cite{rahimi_random_2007}, kernels.

In each scenario, we conduct Adam optimization until convergence with a dynamic learning rate starting from a value of 0.01.
We use a single NVIDIA GeForce 3090-24GB GPU for the experiments. Each trajectory is constituted by triplets $\lbrace \mathbf{x}_i,\dot{\mathbf{x}}_i,\ddot{\mathbf{x}}_i\rbrace_{i=1,\dots,T}$ with $T=1000$.

\Cref{fig:synthetic_results} the different DSs on which we conducted our evaluation.
For each DS we performed 5 training repetitions, each time randomizing training and testing trajectories.
\Cref{tab:synthetic_results} reports the comparison of our approach for first and second order DS against the baseline.
Our method for the first-order DS achieves on average better performance than the baseline.
For the second-order DS, our approach outperforms by a considerable margin the baseline and the first-order DS in RMSE.

\begin{table}[ht]
    \centering
    \resizebox{.45\textwidth}{!}{
\begin{tabular}{l|ccc}
    \toprule
                  & Baseline                              & First DS                              & Second DS                             \\
    Loss          & $6.8\text{e-4} \pm 9.6\text{e-4}$ & $\mathbf{8.7\text{e-5} \pm 2.2\text{e-5}}$ & $1.1\text{e-3} \pm 8.5\text{e-4}$ \\
    Time/Epoch & $0.017 \pm 8.7\text{e-5}$                    & $\mathbf{0.003 \pm 1.9\text{e-4}}$            & $0.011 \pm 1.2\text{e-4}$                    \\
    Train Epochs & $4404.3 \pm 1744.9$ & $\mathbf{4168.3 \pm 797.3}$ & $13257.0 \pm 2280.3$ \\
    \bottomrule
\end{tabular}
}
    \caption{Training results.}
    \label{tab:traininig_results}
\end{table}
\Cref{tab:traininig_results} reports, over 2000 iterations, the training loss and time for each of the tested approaches.
Due to the simplicity of our architecture, our method can be up to $5$ times faster than the baseline for the first-order DS. Despite the increased complexity of the learning problem with respect to the first-order case, our method still manages to be up to two times faster at training time with respect to the baseline.

\section{Robotics Experiments}
\label{sec:robotics_experiments}
\begin{figure}[ht]
    \centering
    \subfloat[]{\includegraphics[width=.22\textwidth]{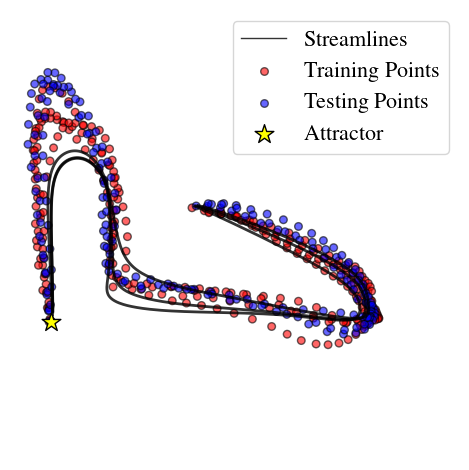}
        \label{fig:robot_learn}}
    \quad
    \subfloat[]{\includegraphics[width=.22\textwidth]{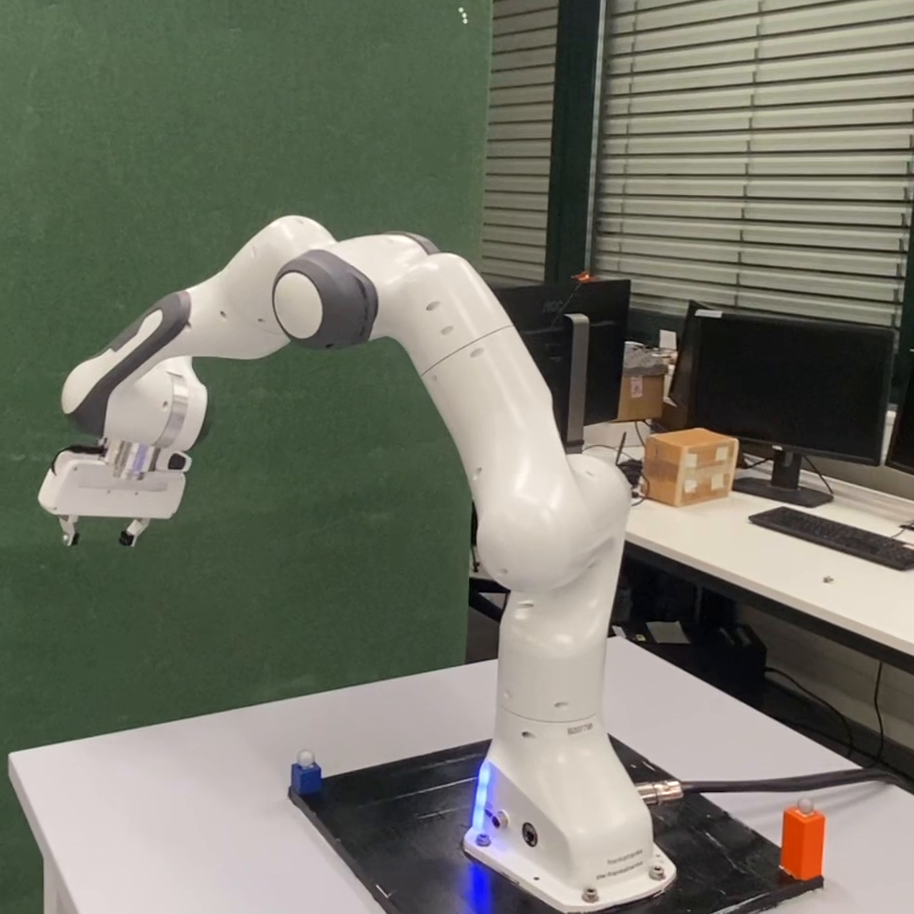}
        \label{fig:robot_setup}}
    \caption{\footnotesize Learning three dimensional Dynamical Systems for Robotic Arm Control.}
    \label{fig:robotic_exp}
\end{figure}

We evaluate our method on learning 3D robotic end-effector motions in real-world settings\footnote{Code to reproduce simulation and real-robot results available at:\\\url{https://github.com/nash169/demo-learn-embedding}}.

\begin{figure*}[t]
    \centering
    \subfloat[]{
        \resizebox{.45\textwidth}{!}{\subimport{../figures/}{control_diagram}}
        \label{fig:control_diagram}
    }
    \hfill
    \subfloat[]{
        \includegraphics[width=.5\textwidth]{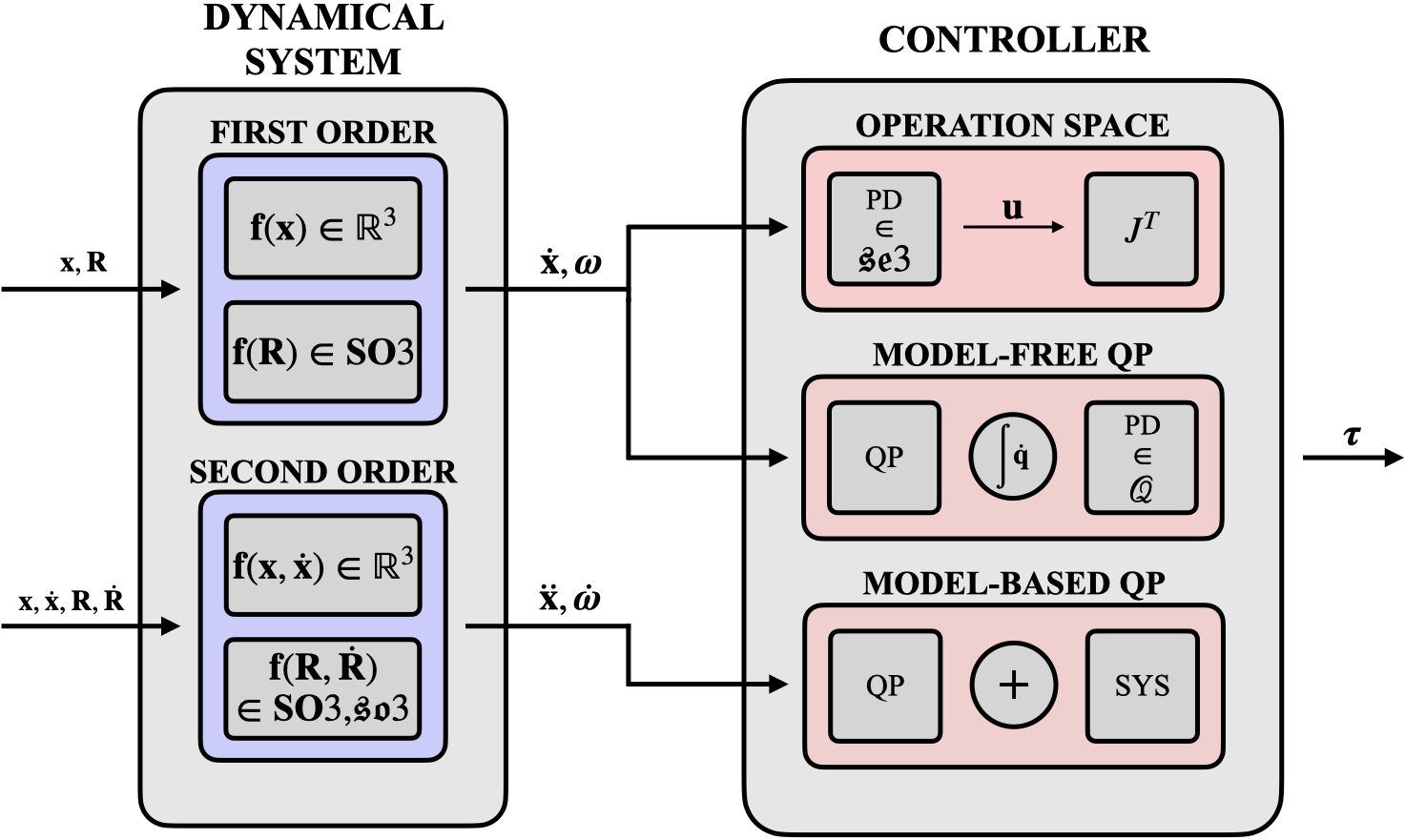}
        \label{fig:block_specs}
    }
    \caption{\footnotesize (a) Control Structure composed by high-level policy generate by the DS and a low-level controller producing torques; (b) depending on the DS deployed torques are generated via operation space control or model-free inverse kinematics QP for first-order DS and model-based inverse dynamics QP for second-order DS.}
\end{figure*}

The robotic platform used consists of the 7-DOF Franka Emila, \Cref{fig:robot_setup}.
We gather 7 samples of the desired task space behavior by manually driving the robot's end-effector.
4 trajectories are selected for training; the remaining trajectories are used as testing set.
\Cref{fig:robot_learn} shows, qualitatively, the 3D learnt DS.
The red dots represent the the sampled observations from the demonstrated trajectories converging towards the attractor represented by the yellow star.
The blue trajectories represents the streamlines obtained by sampling, till convergence, from the learnt 3D DS starting from two different initial locations.

The equations of motion for an articulated robot system can be described as
\begin{equation}
    \m{M}(\mathbf{q})\ddot{\mathbf{q}} + \v{h}(\mathbf{q},\dot{\mathbf{q}}) = \boldsymbol{\tau}_c +  \boldsymbol{\tau}_e\\
    \label{eqn:robot_ds}
\end{equation}
where $\m{M}(\mathbf{q})$ is the inertia matrix, $\v{h}(\mathbf{q},\dot{\mathbf{q}})$ is the sum of gravitational, centrifugal and Coriolis forces.
$\boldsymbol{\tau}_c$ and $\boldsymbol{\tau}_e$ are the vector of controlled and external joint torques, respectively.

Robustness to spatial and temporal perturbation as well as compliancy in the event of human interactions are the desired features of our control strategy.
In this work, we test three different torque-based control strategies.
\Cref{fig:control_diagram} illustrates the generic control strategy where robot configuration space state is fed directly to the controller module while task space information are pre-processed by the learned DS before going inside the controller.
Depending on the DS order we have different control strategies illustrated in details in~\Cref{fig:block_specs}.
The left block shows the two DS tested.
Both the first-order and second-order DSs are split in two submodules, one operating in $\mathbb{R}^3$ and the other one in $SO(3)$, respectively used to control end-effector's position and orientation.
The DS operating in $\mathbb{R}^3$ is learned based on the demonstrated trajectory.
The DS operating in $\text{SO}(3)$ generates a linear DS (critically damped for the second-order space) with the velocities taking place in the Lie Algebra $\mathfrak{so}3$, \cite{sola_micro_2021}.

\paragraph*{Passive Interaction Control}
Model-free control strategy that naturally adapts DS-based control, \cite{kronander_passive_2016}.
It is constitute by a feedback controller with solely the damping term
\begin{equation}
    \boldsymbol{\tau}_c = \m J^T \m D \rbr{\td{\v{x}} - \v{f}(\v{x})} + g(\v q).
\end{equation}
$\v{f}(\v{x})$ is the target velocity generated by the first-order DS, top-left in~\Cref{fig:block_specs}.

\paragraph*{Model-Free Quadratic Programming Control}
In model-free QP control we first find the desired joint velocities as a solution of the optimization problem
\begin{align}
    \min_{\td{\v{q}}, \boldsymbol{\xi}} & \quad \frac{1}{2} \td{\v{q}}^T \m{Q} \td{\v{q}} + \boldsymbol{\xi}^T \m{W} \boldsymbol{\xi} \notag \\
    \text{s.t.}                         & \quad \m J \td{\v q} = \v{f}(\v x) + \boldsymbol{\xi}   \notag                                     \\
                                        & \quad \v{q}^{-} \le \v{q}_{t-1} + \text{dt} \td{\v{q}} \le \v{q}^{+} \notag                        \\
                                        & \quad \td{\v{q}}^- \le \td{\v{q}} \le \td{\v{q}}^+.
\end{align}
The target task space velocity generate by the first-order DS is imposed as relaxed inverse kinematics constraint in the optimization problem.
After prior integration of the desired joint velocities, a feedback controller composed by a proportional and a derivative terms generates the control torques
\begin{equation}
    \boldsymbol{\tau}_c = -\m K \rbr{\v{q} - \v{q}^*} - \m D \v q + g(\v q).
\end{equation}

\paragraph*{Model-Based Quadratic Programming Control}
This control strategy is suited when adopting our learned second-order DS.
In this case the control torques are generated as solution of the optimization problem
\begin{align}
    \min_{\tdd{\v{q}}, \boldsymbol{\tau}, \boldsymbol{\xi}} & \quad \frac{1}{2} \tdd{\v{q}}^T \m{Q} \tdd{\v{q}} + \boldsymbol{\tau}^T \m{R} \boldsymbol{\tau} + \boldsymbol{\xi}^T \m{W} \boldsymbol{\xi} \notag \\
    \text{s.t.}                                             & \quad \m M(\v q) \tdd{\v{q}} = h(\v{q}, \td{\v{q}}) + \boldsymbol{\tau} + \boldsymbol{\xi} \notag                                                  \\
                                                            & \quad \m J \tdd{\v q} = \v{f}(\v x, \td{\v x}) + \boldsymbol{\xi}   \notag                                                                         \\
                                                            & \quad \v{q}^{-} \le \v{q}_{t-1} + \text{dt} \td{\v{q}}_{t-1} + \nicefrac{1}{2} \text{dt}^2 \tdd{\v{q}} \le \v{q}^{+} \notag                        \\
                                                            & \quad \td{\v{q}}^{-} \le \td{\v{q}}_{t-1} + \text{dt} \tdd{\v{q}} \le \td{\v{q}}^{+} \notag                                                        \\
                                                            & \quad \tdd{\v{q}}^-,\boldsymbol{\tau}^- \le \tdd{\v{q}},\boldsymbol{\tau} \le \tdd{\v{q}}^+,\boldsymbol{\tau}^+.
    \label{eqn:qp_second}
\end{align}
In this case we impose track the desired acceleration generated by our second-order DS by imposing a relaxed inverse dynamics constraint in the quadratic optimization problem.
The control torques are
\begin{equation}
    \boldsymbol{\tau}_c = \boldsymbol{\tau}^*,
\end{equation}
where $\boldsymbol{\tau}^*$ is extracted from the solution of the optimization problem in~\Cref{eqn:qp_second}.

\subsection{Trajectory Tracking Evaluation}
In the case of first-order DS-based control, we deploy a standard proportional controller to generate an Euclidean space linear first-order DS that drives the end-effector to the starting point of each testing trajectory. Afterwards, we switch to the learned first-order DS.

When adopting second-order DS-based control, a critically damped proportional and derivative feedback is used to generate a standard Euclidean space linear second-order DS that drives the end-effector to the starting point of each testing trajectory.
Afterwards, we switch to the learned second-order DS in order to perform the desired motion.

\Cref{tab:experiments_results} reports the robotic experiment\footnote{\textsf{beautiful-bullet} simulator available at:\\\url{https://github.com/nash169/beautiful-bullet}\\\textsf{franka-control} control interface for Franka Research 3 available at:\\\url{https://github.com/nash169/franka-control}} results. Compared to model-based QP, Operation Space control and model-free QP yield worse performance in terms of DTWD but achieves much higher control frequency.
The combination of model-based QP and second-order DS demonstrates the best DTWD.
This comes at the cost of a reduced control frequency.
Nevertheless the response time of such control makes it suitable for a large variety of high frequency applications.
\begin{table}[ht]
    \centering
    \resizebox{.48\textwidth}{!}{
\begin{tabular}{lcccc}
    \toprule
                    & \multicolumn{2}{c}{\textbf{$1$st Order DS}} &                                & \textbf{$2$n Order DS}                                \\
    \cline{2-3} \cline{5-5}                                                                                                                                \\
    \textbf{Metric} & \textbf{Operation Space}                    & \textbf{QP Inverse Kinematics} &                        & \textbf{QP Inverse Dynamics} \\
    \midrule
    DTWD            & $2.26 \pm 0.22$                             & $2.57 \pm 0.1$                 &                        & $\mathbf{2.17 \pm 0.25}$     \\
    Frequency [Hz]  & $\mathbf{\approx 1300}$                     & $\mathbf{\approx 1300}$        &                        & $\approx 500$                \\
    \bottomrule
\end{tabular}
}
    \caption{Experimental results.}
    \label{tab:experiments_results}
\end{table}

\section{Conclusion}
\label{sec:conclusion}
In this study, we presented an approach for learning non-linear DS grounded in a purely geometrical framework.
Our approach involves defining a harmonic damped oscillator on a latent manifold.
The inherent non-linearity of the DS is intrinsically captured by the curvature of this manifold so that the chart space representation of the DS's vector field accurately replicates target trajectories.
Our method ensures global asymptotic stability, which is maintained irrespective of the manifold's curvature.
Additionally, our method's explicit embedded manifold's representation grants direct control over the curvature of the space.
This feature is particularly advantageous in integrating the learning of non-linear DS with scenarios involving obstacle avoidance.
Our approach is characterized by relatively minimal constraints.
Specifically, the function approximator is required to learn a multi-scalar function from $\mathbb{R}^d$ to $\mathbb{R}$, maintaining only $C^1$-regularity.
This constraint significantly reduces computational demands during both the training and query phases.

In conventional robotic motion generation, the process typically involves initially planning trajectories at a kinematic level, followed by the development of controllers for accurate trajectory tracking.
Our approach aims at reincorporating dynamics information within LfD framework by integrating two elements: (1) the utilization of high-level policies represented as more expressive second-order DSs, and (2) the application of model-based QP control for efficient one-step ID.

\paragraph*{Limitations \& Future Developments}
We demonstrated the application of learned DS.
We believe that our approach can be adapted for joint space learning without violating joint limits.
This involves initially learning a DS in high-dimensional joint space, followed by the application of local deformations to create energetic barriers, ensuring the robot remains within joint limits.
In order to learn the embedding, we adopted a simple feed-forward network.
Nevertheless, the use of controlled-smoothness kernels like the Matern kernel, coupled with probabilistic embedding, enhances noise robustness.
Specifically, Gaussian Process Regression models can improve precision and query time, making our approach viable for online and adaptive learning.
It is important to note that not every \(d\)-dimensional manifold can be isometrically embedded in a \(d+1\) dimensional Euclidean space.
This limitation constrains the extent of non-linearity that can be effectively learned.
A potential avenue for future research involves exploring embedding strategies for the manifold in a \(d+n\) dimensional Euclidean space.
However, this approach would necessitate the use of \(n\) function approximators, potentially leading to a decrease in model interpretability.
Currently, our methods cannot handle vector fields with limit-cycles or non-zero curl components.
To address the first limitation, one could consider embedding \emph{compact} Riemannian manifolds into higher-dimensional Euclidean spaces.
For the second limitation, involving the manifold topology and the generation of non-zero curl vector fields, extending the theory to the embedding of pseudo-Riemannian manifolds into higher-dimensional Minkowski spaces may offer a solution.

\begin{acks}
    Funding from the European Research Council (ERC) under the European Union's Horizon 2020 research and innovation program, Advanced Grant agreement No 741945, Skill Acquisition in Humans and Robots.
\end{acks}

\bibliographystyle{bibliography/SageH}
\bibliography{bibliography/references,bibliography/references_add}

\appendix

\section{Proofs of Prop.~\ref{prop:embedding} \& Thm.~\ref{thm:stability}}
In this appendix, we provide proofs for Prop.~\ref{prop:embedding} and Thm.~\ref{thm:stability}.

\subsection{Proof of Prop.~\ref{prop:embedding}}
\label{app:proof_prop}
A smooth embedding is an injective \emph{immersion} $f:\mathcal{M} \rightarrow \mathcal{N}$ that is also a \emph{topological embedding}. $f$, in order to be a topological embedding, has to yield a homeomorphism between $\mathcal{M}$ and $f(\mathcal{M})$. Every map that is injective and continuous is an embedding in the topological sense.

Consider the local representative function in Eq.~\ref{eqn:embedding}. Continuity follows directly from imposing $\psi \in C^r(\mathbb{R}^d)$ with $r \ge 1$. Also the injectivity property follows by the construction of the embedding. Let $\mathbf{x} \in \mathbb{R}^d$ and $\tilde{\mathbf{x}} \in \mathbb{R}^d$ two different points on $\mathcal{M}$, expressed in local coordinates, satisfying $\mathbf{\Psi}(\mathbf{x}) = \mathbf{\Psi}(\tilde{\mathbf{x}})$. Therefore
\begin{equation}
  \begin{bmatrix}
    \mathbf{x} \\
    \psi(\mathbf{x})
  \end{bmatrix} =
  \begin{bmatrix}
    \tilde{\mathbf{x}} \\
    \psi(\tilde{\mathbf{x}})
  \end{bmatrix}.
  \label{eqn:injectivity}
\end{equation}
In Eq.~\ref{eqn:injectivity} the equality holds only if $\mathbf{x} = \tilde{\mathbf{x}}$.

$f:\mathcal{M} \rightarrow \mathcal{N}$ is an immersion if $f_*:T_p\mathcal{M} \rightarrow T_p\mathcal{N}$ is injective for every point $p \in \mathcal{M}$; equivalently $\text{rank} (f_*) = \text{dim} (\mathcal{M})$. In order to analyze the rank of the pushforward map $f_*$ we can look at its local coordinates components
\begin{equation}
  f_{*_j}^i = \partial_j(y \circ f \circ x^{-1})^i.
\end{equation}
It is clear from Eq.~\ref{eqn:embedding} that for $i \le \text{dim}(\mathcal{M})$ we have $f_{*_j}^i \equiv \delta^i_j$ where $\delta^i_j$ is the Kronecker symbol. Therefore $\text{rank} (f_*) = \text{dim} (\mathcal{M}) = d$.

\subsection{Proof of Thm.~\ref{thm:stability}}
\label{app:proof_thm}
Stability of the harmonic linearly damped oscillator in Eq.~\ref{eqn:ds_intrinsic} can be proved via Lyapunov's second method for stability. Let $\gamma : I \rightarrow \mathcal{M}$ a curve on the manifold $\mathcal{M}$. We adopt the following energetic function in intrinsic notation
\begin{equation}
  V(\gamma(t), v_{\gamma(t)}) = \frac{1}{2} g_{(\gamma(t))} \left(v_{\gamma(p)},v_{\gamma(t)}\right) + \phi(\gamma(t)),
\end{equation}
where $t \in I$ and $\gamma(t) \equiv p \in \mathcal{M}$. The time derivative, $D_t$, of $V$ is given by
\begin{align}
  D_t V(p, v_{p}) & = \nabla_{v_{p}} V(p, v_{p})                                                             \\
                  & = \frac{1}{2} \nabla_{v_{p}} \left(g_{(p)}(v_{p},v_{p})\right) + \nabla_{v_{p}} \phi(p).
  \label{enq:derivative_lyap}
\end{align}
For the compatibility of the Levi-Civita connection with the metric, $\nabla_{v_{p}} g = 0$, we can simplify Eq.~\ref{enq:derivative_lyap} in
\begin{equation}
  D_t V(p, v_{p}) =  g_{(p)} \left(\nabla_{v_{p}} v_{p},v_{p}\right) + \nabla_{v_{p}} \phi(p).
  \label{enq:derivative_lyap_compatible}
\end{equation}

Eq.~\ref{enq:derivative_lyap_compatible} can be expressed in local coordinates as
\begin{align}
  D_t V(p, v_{p}) & = g_{(p)} \left( -g^{ik} (\partial_i \phi + D_{ij} \dot{x}^j) \frac{\partial}{\partial x^k}, \dot{x}^i \frac{\partial}{\partial x^i} \right) \notag                 \\
                  & + \partial_i \phi \dot{x}^i \notag                                                                                                                                  \\
                  & = g_{(p)} \left(  \frac{\partial}{\partial x^k},  \frac{\partial}{\partial x^i} \right) \left( -g^{ik} (\partial_i \phi + D_{ij} \dot{x}^j)\dot{x}^i \right) \notag \\
                  & + \partial_i \phi \dot{x}^i \notag                                                                                                                                  \\
                  & = -g_{ki}g^{ik} (\partial_i \phi + D_{ij} \dot{x}^j) \dot{x}^i + \partial_i \phi \dot{x}^i
\end{align}
Given the symmetry of the metric tensor $g^{ik} = g^{ki}$ we have
\begin{align}
  D_t V(p, v_{p}) & = -\partial_i \phi \dot{x}^i - D_{ij} \dot{x}^j \dot{x}^i + \partial_i \phi \dot{x}^i \notag \\
                  & = - D_{ij} \dot{x}^j \dot{x}^i
\end{align}
We assumed $D \in \mathcal{S}^d_{++}$. Hence it follows $D \succ 0$ and $D_t V(p, v_{p}) < 0$.

\begin{table*}[t]
  \centering
  \resizebox{.8\textwidth}{!}{
\begin{tabular}{cccccccc}
    \toprule
    \multicolumn{8}{c}{\textbf{MSE - }$\mathbf{1\text{e-}3}$}                                                                                                                                                                                         \\
    \midrule
    \multicolumn{2}{c}{}                                       & \multicolumn{6}{c}{\textbf{NEURONS PER LAYER}}                                                                                                                                       \\
    \cline{3-8}
                                                               &                                                & \multicolumn{1}{|c}{8} & 16                & 32                         & 64                & 128               & 256               \\
    \cline{2-8}
    \multirow{6}{*}{\rotatebox[origin=c]{90}{\textbf{LAYERS}}} & \multicolumn{1}{|c|}{1}                        & 1.935 $\pm$ 2.144      & 0.839 $\pm$ 0.798 & 1.317 $\pm$ 0.652          & 0.632 $\pm$ 0.306 & 1.066 $\pm$ 0.528 & 0.543 $\pm$ 0.314 \\
                                                               & \multicolumn{1}{|c|}{2}                        & 0.209 $\pm$ 0.074      & 1.195 $\pm$ 2.370 & $\mathbf{0.133 \pm 0.020}$ & 0.183 $\pm$ 0.083 & 0.159 $\pm$ 0.096 & 2.092 $\pm$ 2.073 \\
                                                               & \multicolumn{1}{|c|}{3}                        & 0.335 $\pm$ 0.325      & 0.192 $\pm$ 0.061 & 1.093 $\pm$ 1.990          & 0.193 $\pm$ 0.115 & 0.477 $\pm$ 0.480 & 3.507 $\pm$ 3.777 \\
                                                               & \multicolumn{1}{|c|}{4}                        & 0.232 $\pm$ 0.080      & 0.161 $\pm$ 0.077 & $\mathbf{0.121 \pm 0.040}$ & 0.718 $\pm$ 1.282 & 2.094 $\pm$ 1.984 & 1.972 $\pm$ 2.235 \\
                                                               & \multicolumn{1}{|c|}{5}                        & 1.226 $\pm$ 2.111      & 0.844 $\pm$ 1.507 & 1.221 $\pm$ 2.296          & 0.702 $\pm$ 0.766 & 2.564 $\pm$ 1.451 & 2.598 $\pm$ 2.406 \\
                                                               & \multicolumn{1}{|c|}{6}                        & 0.686 $\pm$ 0.554      & 1.191 $\pm$ 2.356 & 2.318 $\pm$ 2.860          & 2.542 $\pm$ 2.516 & 3.887 $\pm$ 1.637 & 7.678 $\pm$ 4.950 \\
    \bottomrule
\end{tabular}
}
  \caption{Ablation Study for the Neural Network function approximator.}
  \label{tab:statistics_results}
\end{table*}

\section{Kernel-Based Space Deformation}
In this appendix we analyze kernel based deformation.
In Sec.~\ref{sec:obstacle_avoidance} we saw how this technique to effectively achieve obstacle avoidance.

\subsection{Derivation of the Metric Tensor}
\label{app:derivation_metric}
The differential of the deformation function in the direction $\mathbf{v}$ is given by
\begin{align}
    d \psi(\mathbf{x}) [\mathbf{v}] & = \text{lim}_{t \rightarrow 0} \frac{\exp{-\frac{\norm{\mathbf{x} + t\mathbf{v} - \bar{\mathbf{x}}}^2}{2 \sigma^2}} - \exp{-\frac{\norm{\mathbf{x} - \bar{\mathbf{x}}}^2}{2 \sigma^2}}}{t} \notag \\
                                    & = k(\mathbf{x}, \bar{\mathbf{x}}) \cdot \text{lim}_{t \rightarrow 0} \frac{1}{t} \left( e^z - 1 \right),
    \label{eqn:differential_kernel}
\end{align}
where $z = -\frac{1}{\sigma^2}(t (\mathbf{x} - \bar{\mathbf{x}})^T \mathbf{v} + t^2 \mathbf{v}^T\mathbf{v})$. Dividing and multiplying \Cref{eqn:differential_kernel} by $z$ and using the property $\text{lim}_{z \rightarrow 0} \left( \frac{e^z - 1}{z} \right) = 1$ we have
\begin{align}
    d \psi(\mathbf{x}) [\mathbf{v}] & = k(\mathbf{x}, \bar{\mathbf{x}}) \cdot \text{lim}_{z \rightarrow 0} \left( \frac{e^z - 1}{z} \right) \cdot \text{lim}_{t \rightarrow 0} \frac{z}{t} \notag              \\
                                    & = k(\mathbf{x}, \bar{\mathbf{x}}) \cdot \text{lim}_{t \rightarrow 0} -\frac{1}{\sigma^2}((\mathbf{x} - \bar{\mathbf{x}})^T \mathbf{v} + t \mathbf{v}^T\mathbf{v}) \notag \\
                                    & = \innerp{-\frac{1}{\sigma^2}(\mathbf{x} - \bar{\mathbf{x}})k(\mathbf{x}, \bar{\mathbf{x}})}{\mathbf{v}} = \innerp{\nabla \bar{\psi(\mathbf{x})}}{\mathbf{v}}.
\end{align}
Via the pull-back of the embedding metric we recover the metric onto the manifold. In case of Euclidean (identity) metric for the ambient space we have
\begin{equation}
    \m{G}(\mathbf{x}) = \mathbf{I} + \frac{1}{\sigma^4}(\mathbf{x} - \bar{\mathbf{x}})(\mathbf{x} - \bar{\mathbf{x}})^T k(\mathbf{x}, \bar{\mathbf{x}})^2.
\end{equation}
The generic sum of kernels formulation is given by
\begin{equation}
    \m{G}(\mathbf{x}) = \mathbf{I} + \frac{1}{\sigma^4}\sum_{i=1}^N \alpha_i(\mathbf{x} - \bar{\mathbf{x}}_i)(\mathbf{x} - \bar{\mathbf{x}}_i)^T k(\mathbf{x}, \bar{\mathbf{x}}_i)^2.
\end{equation}

\subsection{Derivation of the Christoffel Symbols}
\label{app:derivation_christoffel}
Let $\tilde{\v x} = \v x - \bar{\v x}$. The differential of the metric in the direction $\v v$ is given by
\[\label{eqn:metric_derivative}
    d\m{G}(\mathbf{x})[\mathbf{v}] &= \lim_{t \rightarrow 0} \frac{1}{t}
    \left(
    \m{I} + \frac{1}{\sigma^4} \rbr{\tilde{\v x} + t \v v} \rbr{\tilde{\v x} + t \v v}^T k(\tilde{\v x} + t \v v)^2 \right. \notag \\
    & \left. - \m I - \frac{1}{\sigma^4} \tilde{\v x}\tilde{\v x}^T k(\tilde{\v x})^2
    \right) \notag \\
    &= \frac{1}{\sigma^4} \rbr{\v{v}\tilde{\v x}^T + \tilde{\v x}\v{v}^T}k(\tilde{\v x})^2 + \notag \\
    & \underbracket{\lim_{t \rightarrow 0} \frac{1}{\sigma^4 t} \rbr{\tilde{\v x}\tilde{\v x}^T \rbr{k(\tilde{\v x} + t \v x)^2 - k(\tilde{\v x})^2}}}_{\frac{\tilde{\v x}\tilde{\v x}^T}{\sigma^4} d\psi(\tilde{\v x})^2[\v v]} \notag \\
    & = \frac{1}{\sigma^4} \rbr{\v{v}\tilde{\v x}^T + \tilde{\v x}\v{v}^T}k(\tilde{\v x})^2 \notag \\
    & + \frac{1}{\sigma^8}\tilde{\v x}\tilde{\v x}^T\rbr{\v{v}^T \tilde{\v x}\tilde{\v x}^T \v{v}}k(\tilde{\v x})^2.
\]
Regrouping the terms in $\nicefrac{1}{\sigma^4}$ and $k(\tilde{\v x})^2$ we can express the differential of $\m{G}$ as
\[
    d\m{G}(\mathbf{x})[\mathbf{v}] &= \frac{1}{\sigma^4}\left( (\mathbf{v}\tilde{\mathbf{x}}^T + \tilde{\mathbf{x}}\mathbf{v}^T)k(\tilde{\mathbf{x}}) \right. \notag \\
    & \left. + \frac{1}{\sigma^4}\tilde{\v x}\tilde{\v x}^T\rbr{\v{v}^T \tilde{\v x}\tilde{\v x}^T \v{v}} \right)k(\tilde{\mathbf{x}})^2.
\]
The Christoffel symbols of the first---without pre-multiplication for the metric inverse---multiplied by velocity can be derived by permuting the metric derivative as prescribed by~\Cref{eqn:christoffel_second_kind}.
This term is equivalent to what in mechanics is called Coriolis or apparent forces.

\section{Ablation Study}
\label{app:ablation_study}
In order to select the correct model for the neural network used to learn the manifold embedding we performed an ablation study.
In order to asses properly the model's ability of learning the embedding, for the ablation study, we fixed the stiffness matrix to be spherical and we do not optimize for it.
For the second-order DS, the damping matrix is set to be spherical and fixed as well, with diagonal values such that the systems exhibits critically damped behavior in flat space.
In this case the non-linearity is achieve solely via the manifold's curvature.
For different number of layers and neurons within each layer we train each model till convergence.
The training and testing dataset are given by 4 and 3 demonstrated trajectories, respectively.
Each trajectory is composed by 1000 sampled points.
Results are averaged over 10 runs of ADAM optimization on an NVIDIA RTX4090 24GB.

Tab.~\ref{tab:statistics_results} reports the MSE results for different configurations.
As it possible to see from the results, a 2-layers configuration with 32 neurons per layer is enough to achieve good performance with a high level of repeatability.
By increasing the number of layers to 4, we observe an improvement of the performance.
Nevertheless, we did not consider this marginal improvement sufficient to justify the additional overhead in term of computational cost at training and query time.
Therefore, for both the 2D and 3D experiments we opted for a neural network model composed by 2 hidden layers each of the composed by 32 neurons.

\end{document}